\documentclass[]{llncs}

\usepackage[review,year=2024,ID=9808]{eccv}

\usepackage{eccvabbrv}

\usepackage{graphicx}
\usepackage{booktabs}

\usepackage[accsupp]{axessibility}  
\usepackage{soul}

\usepackage[pagebackref,breaklinks,colorlinks,citecolor=eccvblue]{hyperref}

\usepackage{orcidlink}

\begin{document}

\title{DXAI: Explaining Classification by Image Decomposition} 


\author{\authorBlock}


\maketitle

\begin{abstract}
We propose a new way to explain and to visualize neural network classification through a decomposition-based explainable AI (DXAI).
Instead of providing an explanation heatmap, our method yields a decomposition of the image into class-agnostic and class-distinct parts, with respect to the data and chosen classifier. Following a fundamental signal processing paradigm of analysis and synthesis, the original image is the sum of the decomposed parts. We thus obtain a radically different way of explaining classification. The class-agnostic part ideally is composed of all image features which do not posses  class information, where the class-distinct part is its complementary.
This new visualization can be more helpful and informative in certain scenarios, especially when the attributes are dense, global and additive in nature, for instance, when colors or textures are essential for class distinction. Code is available at \url{https://github.com/dxai2024/dxai}.
\end{abstract}

\section{Introduction}
\label{sec:intro}


\begin{figure*}[htb]
\centering
\includegraphics[width=0.9\textwidth]{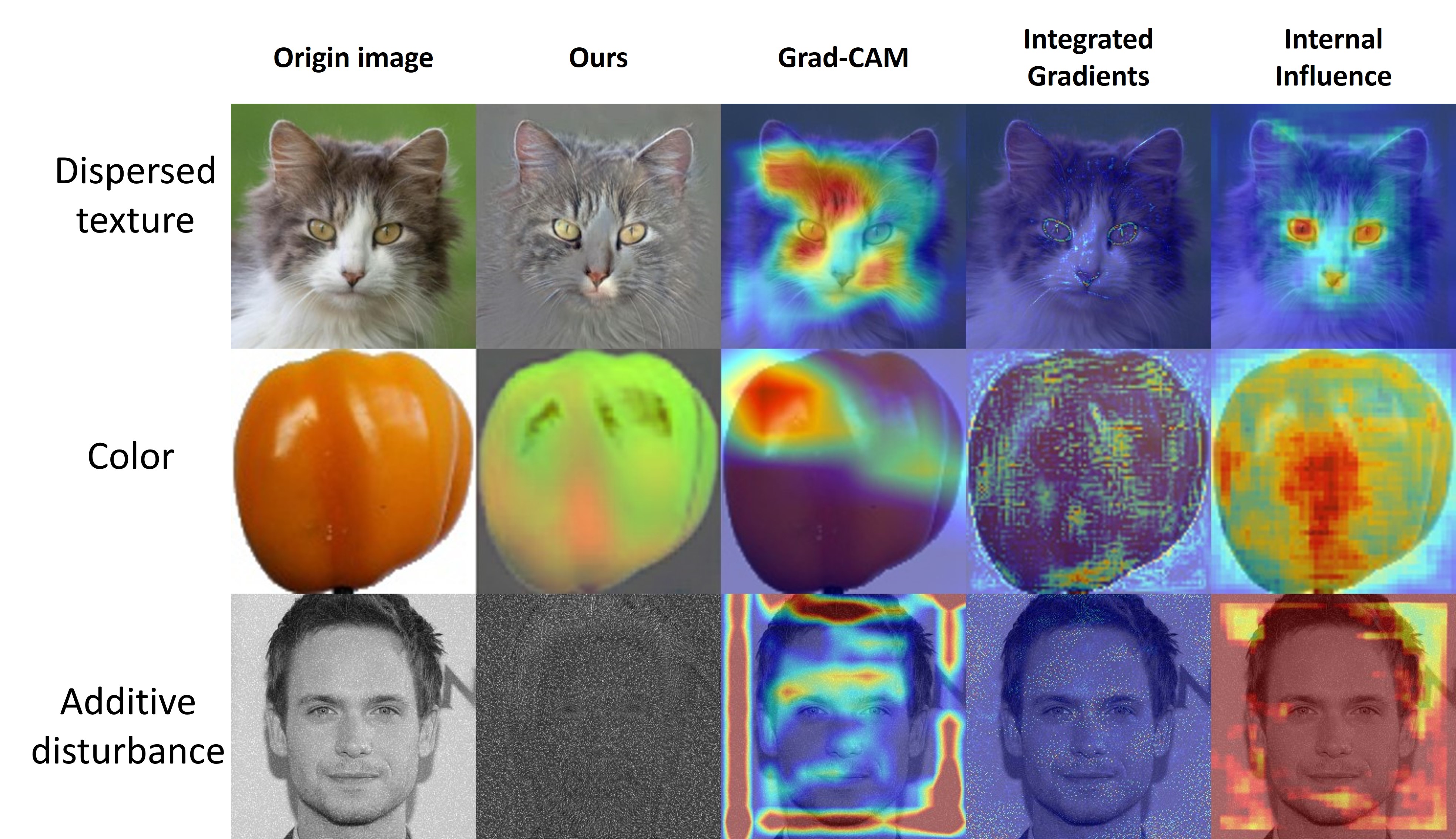}
\caption{Our method vs. heatmaps, illustrating three scenarios where heatmaps are less informative: 1) Many details spread across large portions of the image are helpful for accurate classification (top row), heatmaps show only partial relevant information. 2) Distinguishing between types of peppers that differ mainly by color (2nd row). 3) Detecting additive statistical disturbance (a class of clean images and a class of images with noise). Since the contribution is global --  heatmaps face difficulties  explaining the reason for classification.}
\label{fig:heatmaps_compare}
\end{figure*}

Understanding classification reasoning of neural networks is of paramount importance. It can lead to better understanding of the classificaction process, help in debugging and validation and also serve as an additional informative output for the user at inference.
Hence, the research on the topic of explainable artificial intelligence (XAI) is extensive \cite{dwivedi2023explainable, kamakshi2023explainable, alicioglu2022survey, ghassemi2021false}.

\begin{figure}[htb]
\centering
\includegraphics[width=0.7\textwidth]{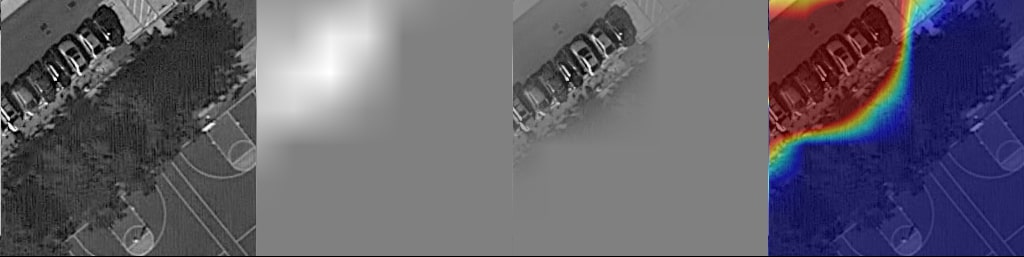}
\caption{Three visualizations using a heatmap. Highlighting cars in an aerial image (left). Generated by Grad-CAM \cite{selvaraju2017grad}.
}
\label{fig:Grad-CAM displaying}
\end{figure}

The most common way to show the explanation of the network in image classification is by producing an \emph{explanation heatmap}. This map is a per-pixel indication of the relevance of that pixel to the final classification decision of the network (see Fig. \ref{fig:Grad-CAM displaying}). In this visualization, the larger the value - the more relevant this pixel is to the classification. These maps can be at some lower resolution, where visualization is done by upsampling, or by using superpixels. In addition, some methods give also negative values, indicating pixels which reduce the confidence for the final classification. The heatmaps themsleves do not resemble actual images and to understand the role of the pixels in a heatmap common practices are either to show the input image and the heatmap side by side, to overlay the heatmap on the image, or (less commonly) to show an image where the brightness of the pixels are weighted by the (normalized) heatmap, \cref{fig:Grad-CAM displaying}.

This visual explanation method is mostly adequate when the explanation is spatially \emph{sparse}. That is, there are just a few small regions in the image which contribute mostly to the classification. However, there are many classification problems in which the explanations are \emph{dense} in the image domain. This can happen in several scenarios, for instance:
\begin{enumerate}
    \item The object to be classified spans a large portion of the image domain and contains many diverse features, all contributing to the final classification.
    \item A main feature contributing to the classification is color change, which appears throughout the image.
    \item The class distinction is based on some global disturbance or statistical change, which spans the entire image domain.
\end{enumerate}
In such scenarios, a heatmap is much less informative.
Depending on the algorithm, it is either too focused on just a small portion of highly dominant features or it shows uniform large areas spanning most of the image. See some examples in \cref{fig:heatmaps_compare}. In both cases, we lack a clear constructive explanation for the network's decision.
Furthermore, the basis for using explanations through heatmaps implicitly assumes that some pixels are crucial for the solution while others are either unimportant or only partially important. We argue that there are scenarios where each pixel contains both class identity information and neutral data, necessitating the use of a different method.

\begin{figure}[htb]
\begin{subfigure}{0.45\textwidth}
\centering
\includegraphics[width=0.9\linewidth]{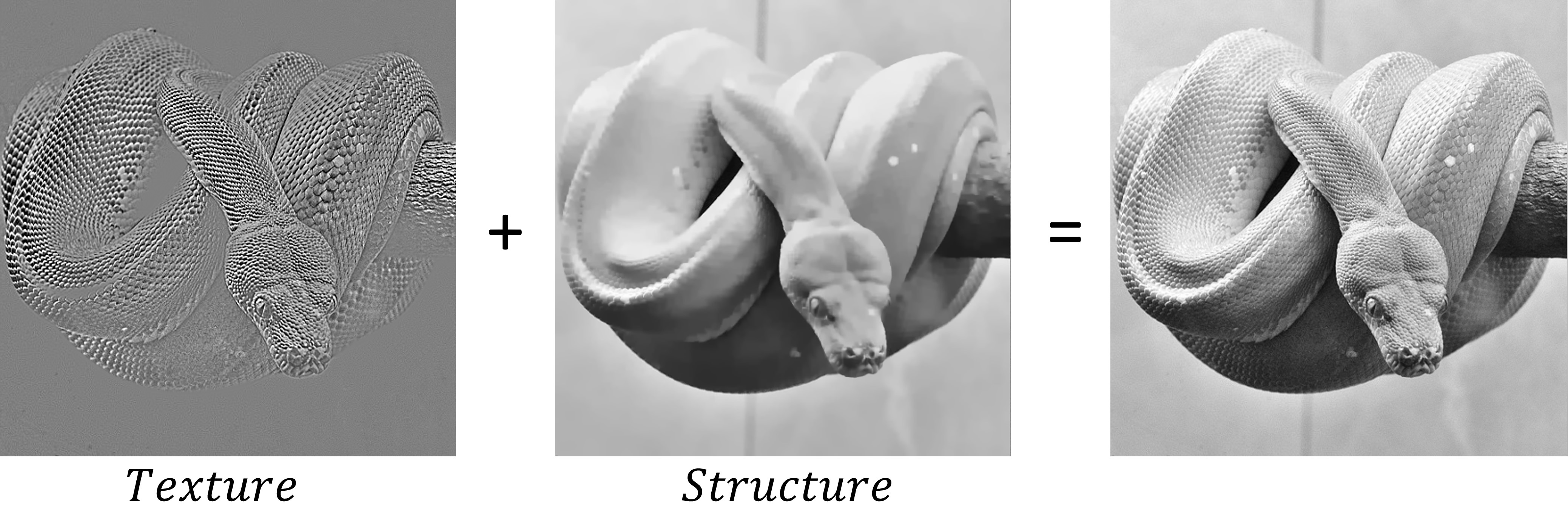}
\includegraphics[width=0.9\linewidth]{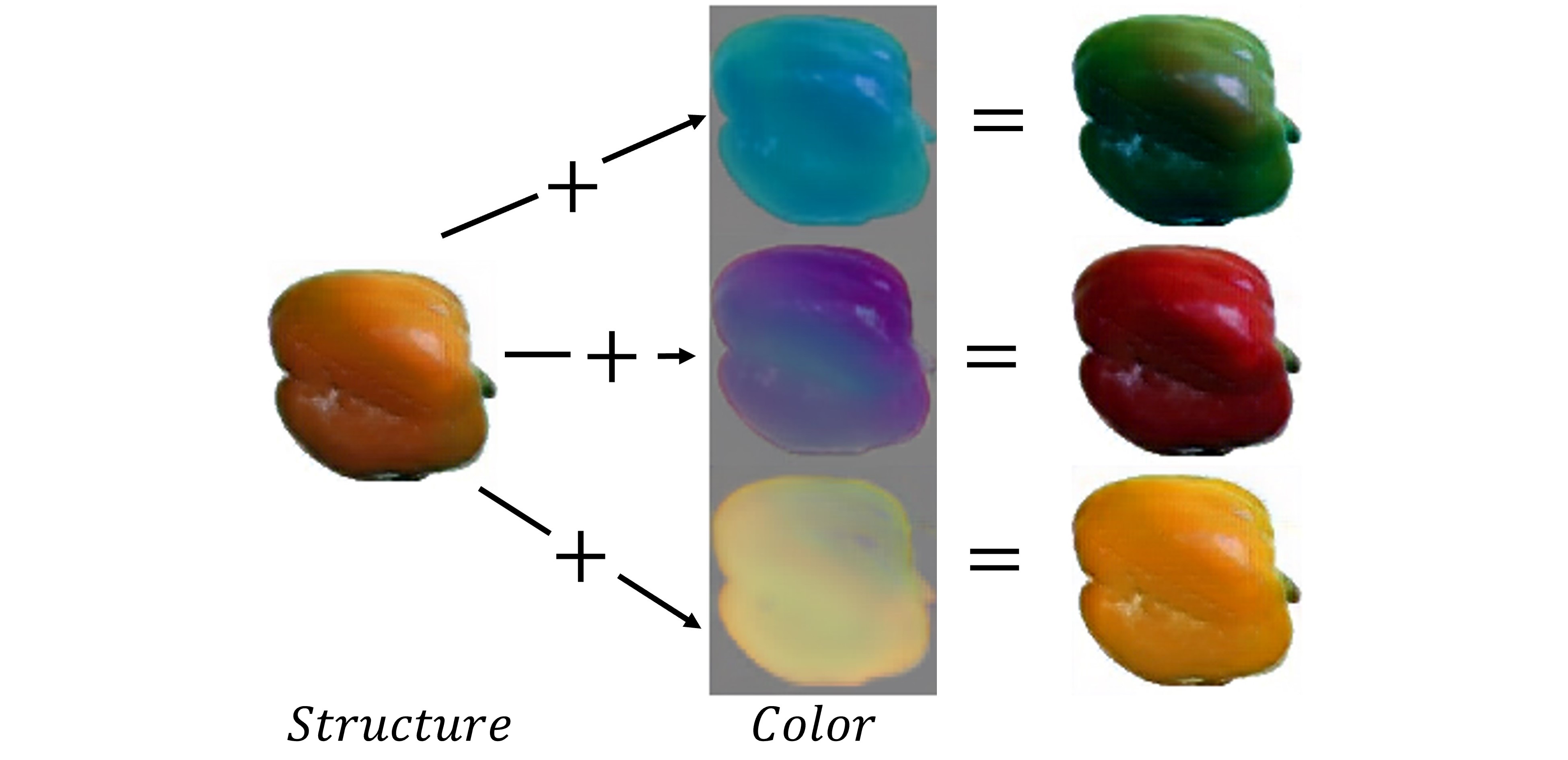}
\caption{\textbf{Additive decompositions.} The top image illustrates texture-structure decomposition  \cite{xu2020structure}. Our explanation is similarly additive, for peppers classification we implicitly obtain a structure-color decomposition.}
\label{fig:additive_decomposition}
\end{subfigure}
\begin{subfigure}{0.45\textwidth}
\centering/
\includegraphics[width=0.9\linewidth]{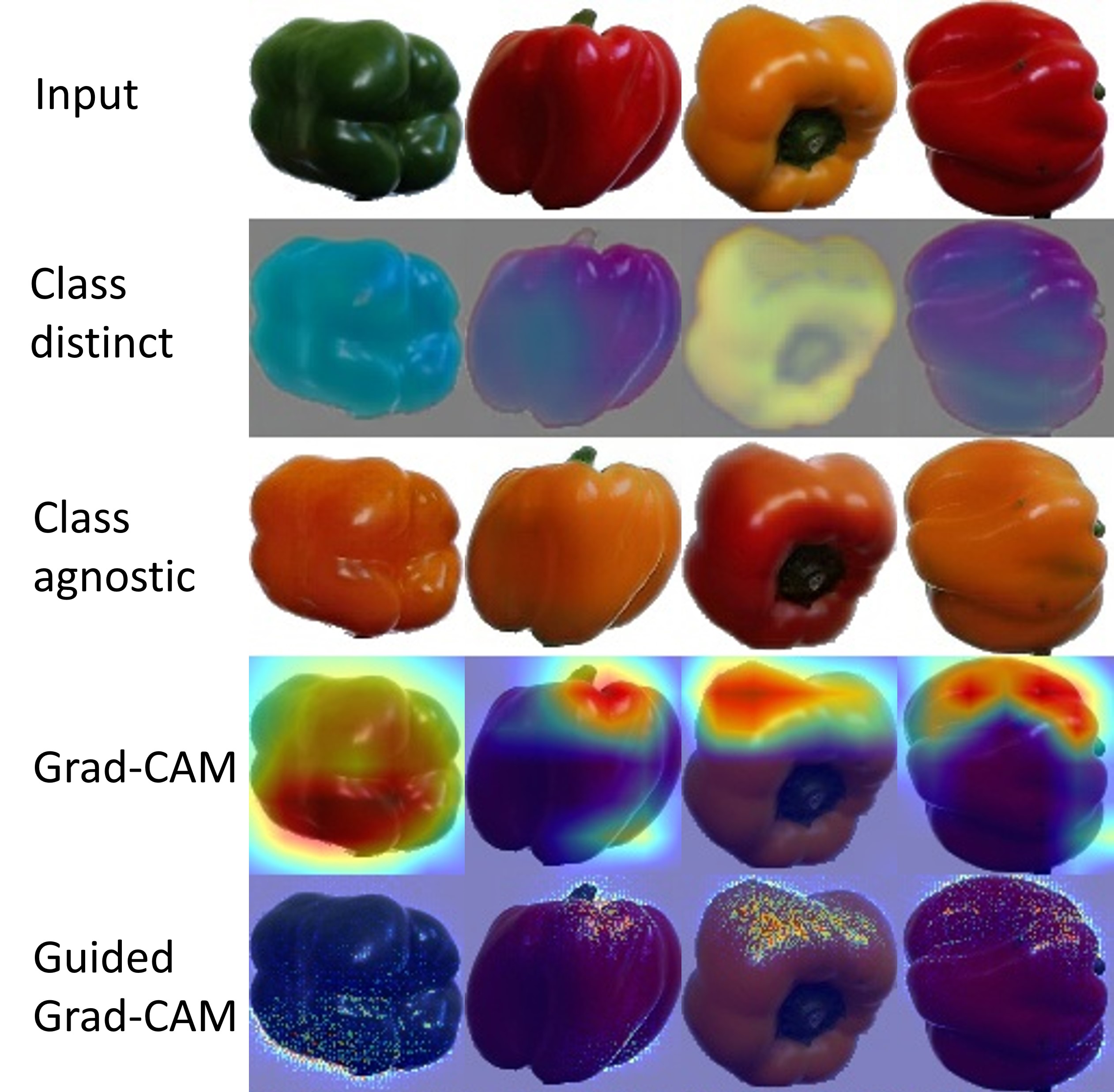}
\caption{Peppers \cite{thi2019fruits}: Class-distinct and Class-agnostic parts. The dataset includes four classes of peppers of different color.}
\label{fig:peppers}
\end{subfigure}

\caption{An image can often be decomposed into meaningful additive parts, such as in classical structure-texture decomposition (top-left). This example shows how it allows to explain well color-based reasoning.}
\end{figure}

Our approach assumes a membership logic, such that each region is potentially a superposition of image features  common to many classes and ones which are class-specific. 
Image decomposition is a long standing paradigm in classical image processing.
It is based on the notion that images (and signals in general) can often be well decomposed into meaningful additive parts.
Given a 1D signal $x$ of length $N$ and some orthonormal basis of $N$ elements $\{ \phi_i \}$, $i=1,..\, N$, one can express $x$ as
$$ x = \sum\limits_{i=1}^N \alpha_i \phi_i, $$
with $\alpha_i = \langle x,\phi_i \rangle$, where $\langle \cdot ,\cdot \rangle$ denotes inner-product. Some well known examples which fall into this category are Fourier 
and wavelet basis \cite{mallat1989theory,daubechies1992ten}.
In overcomplete representations \cite{donoho2005stable,aharon2006k} a common practice is to decompose a signal into only a few $K\ll N $ elements (sparse representation) of an overcomplete dictionary $\{ \psi_i \}$, $i=1,.. D \gg N$, allowing some small error $e$, 
$$ x = \sum\limits_{i=1}^K \alpha_i \psi_i + e. $$
There are many other studies attempting to decompose an image into meaningful additive parts. For example, image decomposition into structure and texture $x = \psi_{Structure}+\psi_{Texture} (+ e)$ (with possible noise or small error)  
\cite{xu2020structure,aujol2006structure,aujol2005dual,fadili2009image}. See example in \cref{fig:additive_decomposition}.

In this work we propose to express XAI by the following image decomposition:
\begin{equation}
    \label{eq:xai_decomp}
    x = \psi_{Agnostic} + \psi_{Distinct},
\end{equation}
where $\psi_{Agnostic}$ is the \emph{class agnostic} part of the image, which ideally does not entail information about the class, and $\psi_{Distinct}$ is the \emph{class distinct} part which holds the discriminative information, allowing the classifier to obtain distinction from other classes. 
We use style transfer based on generative AI to accomplish this. We show this way of explaining classification brings new computational and visualization tools, which, for some cases, are much more natural and informative. Our main contributions are as follows:
\begin{enumerate}
    \item We present a detailed computational framework to estimate Eq. \eqref{eq:xai_decomp}, for a given classifier, training set and classification task. The decomposition is of high resolution, allowing to portray well fine and delicate details. 
    \item We show, for the first time, class agnostic images, based on decomposition. This provides new information and insights on the classification problem.
    \item The method is fast since results are produced at inference time of generative models   (no gradients are computed).
    \item We provide extensive examples and experimental data showing the advantages (and limitations) of the method, compared to heatmaps.
\end{enumerate}

\section{Related Work}
\label{sec:related}

In this work, we focus on visually explaining image classification. Specifically, we concentrate on algorithms that, given an image classification result, highlight the areas contributing to the classification  \cite{selvaraju2017grad, ribeiro2016should,lundberg2017unified,binder2016layer,sundararajan2017axiomatic,vermeire2022explainable, leino2018influence, komorowski2023towards}. As previously discussed, high values indicate significance, while values near 0 are of negligible contribution.
%
A leading approach is backpropagation-based methods \cite{selvaraju2017grad,binder2016layer, sundararajan2017axiomatic}. It involves tracing the classifier's solution backward through the model's layers to measure the contribution of each layer to the subsequent one. These methods encompass gradient-based and attribution-based techniques.
Perturbation-based methods \cite{lundberg2017unified,ribeiro2016should, vermeire2022explainable} evaluate the impact on the output of changes in the input. Changes leading to strong output variations are deemed important.

Attention-based methods \cite{komorowski2023towards, gkartzonika2022learning, xue2022protopformer} identify relationships within the input to discern important image characteristics. These methods often require specific classifier architectures.
Other methods estimate the uncertainty in the solution and use this to find explanations with high confidence \cite{marx2023but, zhang2022explainable}.
Generative models are also used to explain differences between classes. Many of the methods recently try to provide counterfactual explanations \cite{jeanneret2022diffusion, mertes2022ganterfactual}. However, it appears to be difficult to use these explanations in order to produce a map of clearly highlighted differences.
Current XAI methods typically suffer from at least one of the following drawbacks:
\begin{enumerate}
\item {\bf Resolution.} Low resolution that hinders fine detail distinction, usually caused by calculating importance in spatially coarse internal layers.
\item {\bf Compute.} Extended runtime or high memory consumption due to gradient or attribution calculations across multiple layers or using numerous perturbation iterations.
\item {\bf Architecture.} Certain solutions impose architectural constraints, such as specific activation layers or even a dedicated architecture designed solely for XAI.
\item {\bf Single channel.} Typically, these algorithms produce grayscale images representing the importance of each pixel, lacking color distinction (multi-channel data in general). This can lead to inadequate explanations, as seen, for example, in our peppers dataset experiment (\cref{fig:peppers}).
\end{enumerate}
Our proposed approach comprehensively addresses these challenges, as detailed below.

\section{Method}
\label{sec:method}

\begin{figure}[htb]
\centering
\includegraphics[width=0.85\textwidth]{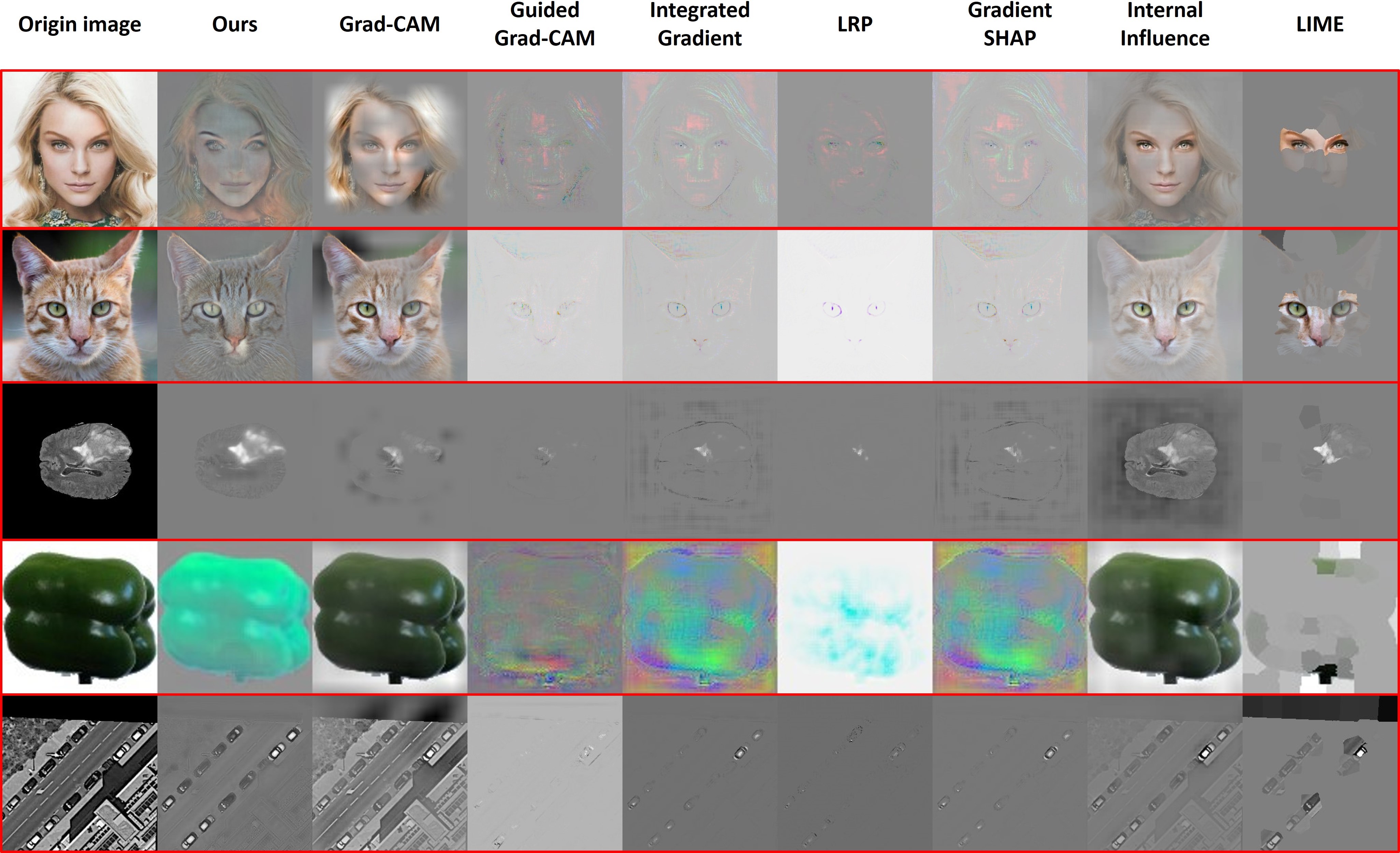}  
\caption{Class-distinct component $\psi_{Distinct}$ by our DXAI algorithm and by heatmap manipulation, in which the input image is weighted by the normalized heatmap (see details in \cref{sec:exp}). We obtain high resolution, dense multi-channel explanations.}
\label{fig:xai_compare}
\end{figure}
We aim at partitioning the image to a component which is neutral for the classifier (agnostic) and to a component which explains the classification (distinct). We provide some definitions and formulate the DXAI problem. This is a very hard nonconvex minimization problem. In the second part we present an architecture and a training procedure to obtain an approximate solution. 


\subsection{The DXAI problem}
We first introduce the basic setting and notations.
We assume a multiclass supervised learning setting.
Let $X$ be the space of input images and $x \in X$ be an input image belonging to one of $c$ classes. Let $y\in Y$ be a class label $y\in\{0,c-1\}$. The training set consists of $M$ pairs $(x_i,y_i) \in X \times Y$, $i=1,..\,,M$.
Let $x^y$ denote an image $x$ belonging to class $y$. An image classifier $C$ is given, which provides a class probability vector $p$. 

We would like to extract the relevant features, and only them, which contribute to the classification decision by the network. An additive contribution, as in  Eq. \eqref{eq:xai_decomp}, is assumed.
We want to have a reference image, which is neutral in terms of classification (for a given classifier), using the following definition:
\begin{definition}[Class Agnostic]
\label{def:agnostic}
    For an image $\psi \in X$, given $c$ classes of images and a classifier $C$ providing a vector $p(\psi)\in \mathbb{R}^c$ of class probabilities, $\psi$ is class agnostic if the probability is uniform, that is  $p_i(\psi)=\frac{1}{c}$, $\forall i=1,.. \,c$, where $p_i$ is the $i$'s entry of $p$.
\end{definition}
We denote by $X_A \subset X$ the space of all $\psi \in X$ which
are class agnostic, as defined above.
Given an image $x\in X$ with class probability vector $p(x)$, and a classifier $C$, we would like to solve the following minimization problem:
\begin{equation}
\label{eq:dxai_problem}
    \min_{\psi \in X_A}d(x,\psi),
\end{equation}
where $d(\cdot, \cdot)$ is some distance measure. 
For instance, our implementation uses a combination of $L^1$ and $L^2$ norms for this distance.
We refer to Eq. \eqref{eq:dxai_problem} as the DXAI (decomposition-based XAI) problem.
We can now define the class distinct and class agnostic parts of an image $x$.
\begin{definition}[Class Distinct / Agnostic parts of  $x$]
\label{def:d_a_x}
    Let $x\in X$ be an image with class probability vector $p(x)$ provided by the classifier $C$.
    Let $\psi^*$ be a minimizer of \eqref{eq:dxai_problem}. Then $\psi_{Agnostic} = \psi^*$ is a class agnostic part of image $x$ and $\psi_{Distinct} = x - \psi^*$ is a class distinct part.    
\end{definition}
In words, for a given image $x$, we find the closest image (with respect to $d$) which is neutral, in terms of class (agnostic part). The difference between $x$ and this neutral image explains the probability vector $p(x)$ and the reason, according to the classifier, why it deviates from neutrality. 
We do not assume a unique solution.
The pair $\{\psi_{Agnostic}, \psi_{Distinct}\}$ provide detailed and dense class explanation. 
Obtaining an approximate agnostic part may be computed for each image directly using optimization techniques. For example, by minimizing a loss that requires minimal distance in terms of KL divergence between the distribution generated by the classifier $C$ and a uniform distribution. However, the distinction component is not 
semantically viable, it is mostly based on out-of-distribution features (which resemble noise, see the example in \cref{ablation}).
Obtaining a solution that captures semantic characteristics is more challenging. We propose to solve this using generative models.

\subsection{Approximate solution for the DXAI problem}
Given an image $x$ classified to class $y$ by the classifier $C$, we want to approximate a pair
$\{\psi_{Agnostic},\psi_{Distinct}\}$.
We use generative models for the decomposition. 
We leverage style transfer as the primary tool for discerning inter-class differences and generating class-explanations. 
Our implementation uses style-transfer GANs \cite{choi2020starganv2}. 
Following \cite{brokman2022analysis}, which shows better descriptive capacity of branched GANs, we use several generators for the agnostic part, which usually contains most of the image data. 
We assume to have $n$ style transfer generators 
$G_i(x^y,s_y)$, which take as inputs an image $x^y$ and a style vector $s_y$. The style vector is generated by a dedicated mapping network, as detailed in \cite{choi2020starganv2}. This network takes the class as input and produces the corresponding style representation.
Given an image $x^y$ of class $y$ we approximate it by the following decomposition into $n$ components (branches):
\begin{equation}
\hat{x}^y =\psi_1^y+\sum\limits_{i=2}^n \psi_i^y,
\label{eq: x^y decomposition}
\end{equation}
where $\hat{x}^y \approx x^y$ and $\psi_i^y=G_i(x^y,s_y)$.
We assign  
\begin{equation}
\label{eq:psi_d_psi_a}
    \psi_{Distinct} = \psi_1^y; \quad \psi_{Agnostic} = \sum\limits_{i=2}^n \psi_i^y.
\end{equation}
Our aim is to transform an image from class $y$ into an image representative of target class $\tilde{y}$. Successful style transfer requires the identification and modification of distinct class-specific characteristics. Style transfer is accomplished by using the style vector associated with class $\tilde{y}$ in the generators. That it, $\psi_i^{\tilde{y}}=G_i(x^y,s_{\tilde{y}})$.

In addition, our architecture incorporates a multi-head discriminator ($D$), a versatile component that serves dual roles within our framework. This discriminator takes an image as input, producing a vector whose length corresponds to the number of classes. Each element in this vector is a ``grade'', reflecting the authenticity of the input image concerning the class it represents. Beyond its role as a discriminator, $D$  also serves as a classifier, effectively classifying the input image by selecting the class with the highest ``grade'' using argmax of its output vector. 

Lastly, we have a pre-trained classifier ($C$), for which DXAI is computed. In this context, the generators aim to deceive the classifier by aligning their outputs with the intended class representation. Here, our multi-head discriminator's primary function is to evaluate the extent to which the generators successfully mislead the pre-trained classifier while ensuring the overall quality and realism of the generated images.

\subsection{Training}
\begin{figure}[htb]
\centering
\includegraphics[width=0.8\textwidth]{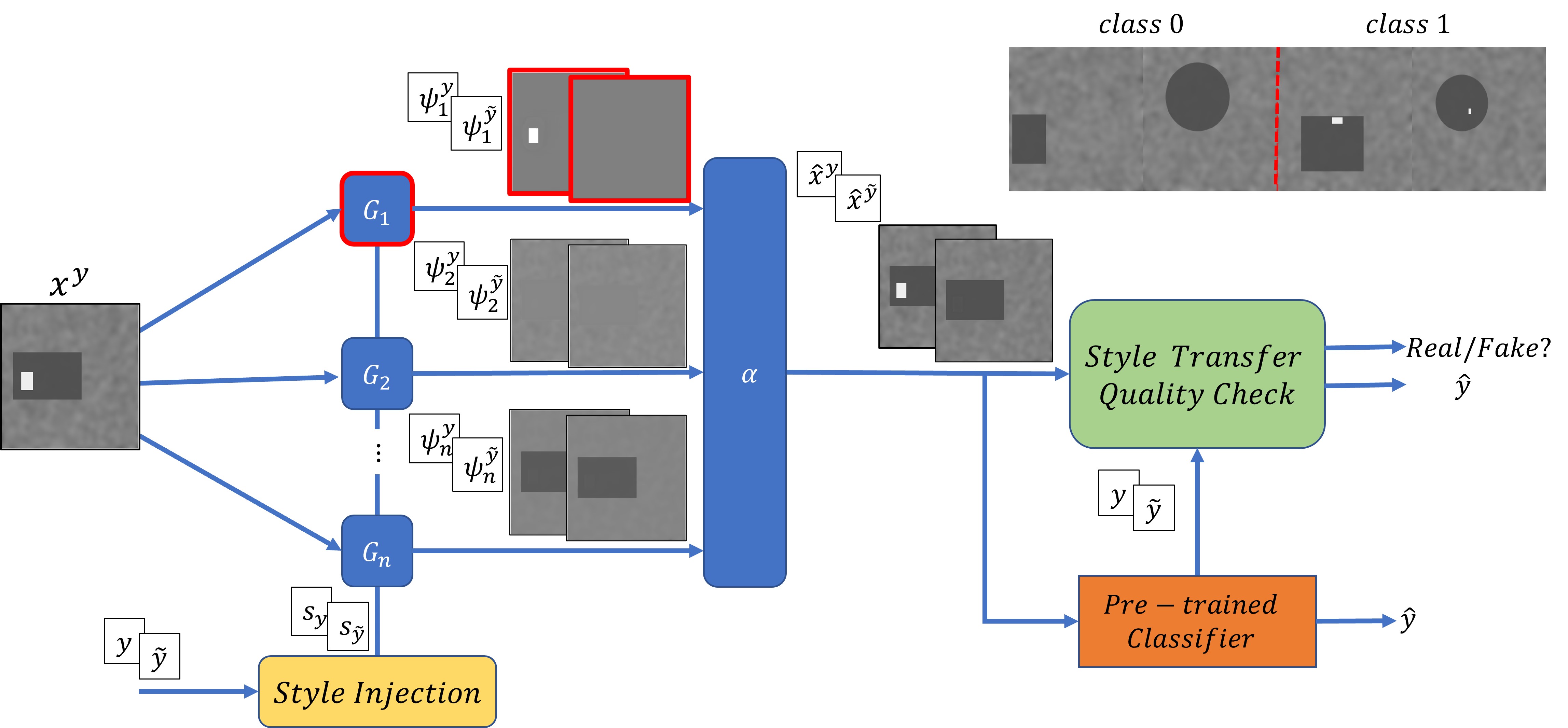}
\caption{{\bf Training process diagram.} The class distinct part (existence/absence of white rectangle) is in the first branch (top, in red), whereas the class agnostic components, which belong to both classes, are generated by subsequent branches. The $\alpha$-blended generation is a major mechanism controlling the training. 
}
\label{fig:training_process_without_classifier}
\end{figure}

We list below the essential training procedure and losses. 
See \cref{fig:training_process_without_classifier} for an overall
diagram of the general architecture, consisting of several style-transfer generators, our proposed $\alpha$-blending procedure, style and quality discriminator and classifier. 
More details, including ablation studies, appear in the supplementary material.  

{\bf New $\alpha$-blended generation.}
In order for the first channel to contain class distinct information, we propose a novel $\alpha$-blending mechanism. 
For each batch, a random vector $\alpha$ of length \(n-1\) is drawn, where each element is uniformly distributed in the range $[0,1]$. Two images are then generated during training as follows:
\begin{equation}
\begin{aligned}
    \hat{x}^y=\psi_1^y+\sum\limits_{i=2}^n \alpha_{i-1}\psi_i^y+\sum\limits_{i=2}^n (1-\alpha_{i-1})\psi_i^{\tilde{y}}, \\
    \hat{x}^{\tilde{y}}=\psi_1^{\tilde{y}}+\sum\limits_{i=2}^n \alpha_{i-1}\psi_i^{\tilde{y}}+\sum\limits_{i=2}^n (1-\alpha_{i-1})\psi_i^y,
\end{aligned}
\label{eq:alpha_comb}
\end{equation}
where $y$ is the class of the input image and
$\tilde{y}$ represents a random alternative class, $\tilde{y} \ne y$.
The proposed method encourages the generators to generate identical images for both components in the sum $\psi_i^y \approx \psi_i^{\tilde{y}}$, $i=2,..\,,n$, and thus to isolate the distinction between the classes to $\psi_1$. In the ideal case, where the components in the sum are identical and the distinction is only in $\psi_1$ we converge to \cref{eq: x^y decomposition}.
The proposed $\alpha$-blending method allows a stable and effective training. We note that other alternatives,
such as attempting to use norm-based losses, e.g. 
$\|\psi_i^y - \psi_i^{\tilde{y}}\|$, often yield degenerate solutions, with $\psi_i^y \approx 0$.




{\bf Classification loss.}
Since a pre-trained classifier is integrated into our system, there is no need to further train it on authentic images. Instead, we leverage its classification and attempt to explain it. We enable the generators to produce images that correspond to the classifier's predictions through the following loss function:
\begin{equation}
L_{class-fake} = CrossEntropy\Bigl(C\Bigl(G(x^y, s_{y_{trg}})\Bigl), y_{trg}\Bigl).
\label{eq:loss_class_fake}
\end{equation}
In our GAN-based model,  the discriminator is used also as a classifier, in addition to its classical role. This is in order to distinguish between real and fake images in each class. We use a Kullback-Leibler divergence  loss between the classification output of the discriminator and that of the pre-trained classifier $C$. This promotes having a high value in the discriminator output only for images which appear real and fit the correct class.

{\bf Reconstruction loss.}

Our generated image $\hat{x}^y$ is only an approximation of $x$, see Eqs.
\eqref{eq: x^y decomposition}, \eqref{eq:alpha_comb}. To obtain a good 
approximation, $\hat{x}^y \approx x$, we use a fidelity measure, based on $L^1$ and $L^2$ norms,
\begin{equation}
d(u,v) = \|u-v\|_{L^1}+\|u-v\|_{L^2},
\label{eq:d}
\end{equation}
to penalize small and large changes.
We would also like the style transferred class to be similar to the input image. Thus, the reconstruction loss is with respect to the generated images of both classes,
\begin{equation}
\label{eq:L_rec}
L_{rec} = d(x,\hat{x}^y) + d(x,\hat{x}^{\tilde{y}}),
\end{equation}
where $\hat{x}^y$, $\hat{x}^{\tilde{y}}$ are given in \eqref{eq:alpha_comb}.


{\bf DXAI and zero distinction.}
Let us see how the training process and losses above approximate the DXAI problem. 
%
The reconstruction loss promotes $x \approx \hat{x}^y \approx \hat{x}^{\tilde{y}}$ .
The classification loss promotes that $\hat{x}^{\tilde{y}}$
belongs to class $\tilde{y} \ne y$. Thus, the shared components $\psi_i^y \approx \psi_i^{\tilde{y}}$, $i=2,..\, n$, should belong to the class-agnostic part. The effects of branch specialization, as shown in \cite{brokman2022analysis}, encourage
each $\psi_i$ to contain different image characteristics.
Since class $\tilde{y}$ is random, following  \eqref{eq:psi_d_psi_a}, we get that $\psi_{Agnostic}$ is not committed to any specific class.

\begin{figure}[htb]
\centering
\includegraphics[width=0.8\textwidth]{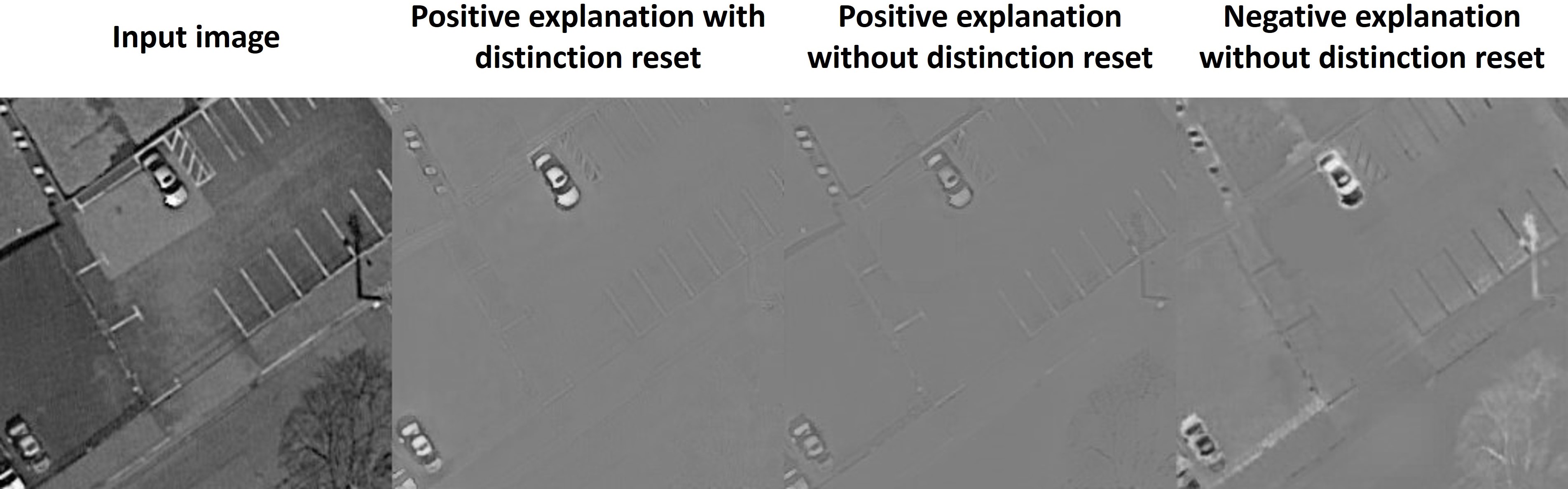}
\caption{\textbf{Effect of resetting the distinction branch.} This reset yields enhanced and cleaner class explanation (cars), in contrast to less prominent and possibly negative explanation otherwise.}
\label{fig:zero_distinction_effect}
\end{figure}

In addition, we chose to set $\psi_1^{\tilde{y}}=0$ in \cref{eq:alpha_comb}.
This is in line with the DXAI formulation, \cref{eq:dxai_problem}.
We would like to choose the agnostic part which is closest to
the input image. We explain below additional benefits of this setting.
Since our algorithm is of additive nature, it can offer two types of explanations. 
The more intuitive approach is to highlight unique class features positively, effectively adding distinctiveness to an image with neutral attributes. This ensures the appearance of the differences in the distinction map. Alternatively, it is possible also to \emph{subtract} distinct features. 
Negative explanations are less preferred, since they are less intuitive for class explanation. For instance, when our classifier predicts an image contains cars, we prefer to receive an image of cars in $\psi_{Distinct}$, rather than a subtraction when it predicts the absence of cars. Setting the distinction of the alternative class to zero diminishes negative explanations.
Moreover, this promotes the full network
to produce as realistic images as possible by the alternative 
generators, reducing spurious features and undesired details in the class-distinct component, see \cref{fig:zero_distinction_effect}.

The total loss is a weighted sum of the mentioned losses,  in addition to the standard adversarial loss of GAN architecture \cite{choi2020starganv2}.
Inference is performed by providing an image $x$ and a class $y$ (the inference diagram and additional training and inference details are in the supplementary).  

\section{Experiments}
\label{sec:exp}


We show here various experiments examining the validity of our algorithm and the applicability of using the proposed method to infer useful classification explanations for various diverse data sets. During the experiments we used three different classifiers. Details about their architecture appear in the supplementary.

\begin{figure}[htb]
\centering
\includegraphics[width=0.9\textwidth]{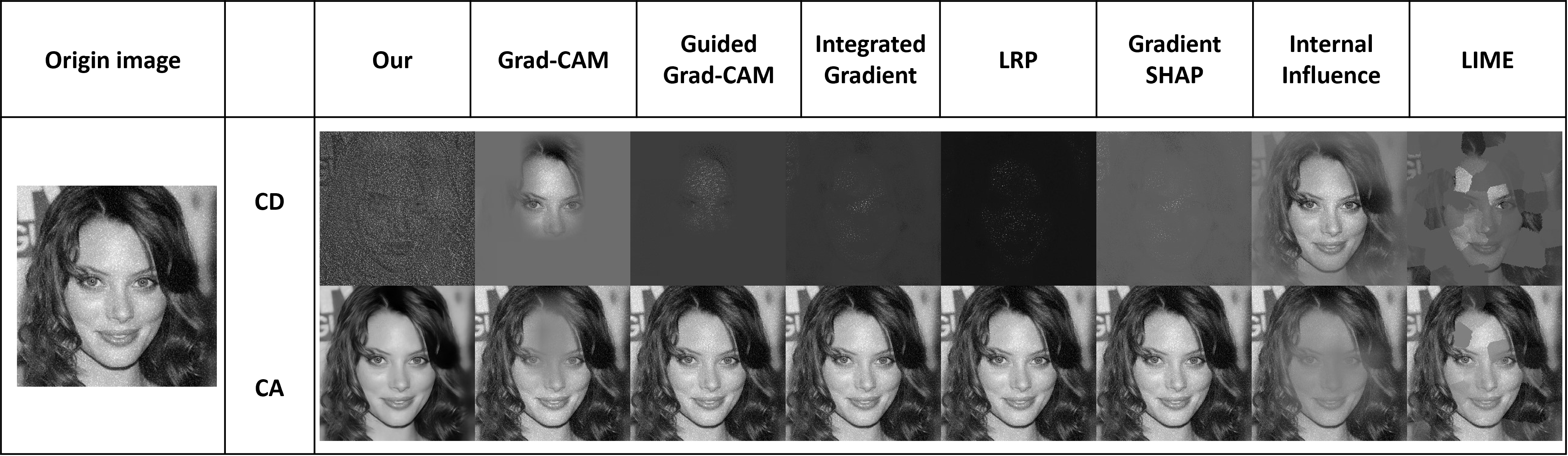}
\caption[Noise distinction results]{{\bf Noise distinction.} One class contains clean images (face images from Celeb-A) and the other class contains noisy images (additive white Gaussian noise). In this case, standard XAI algorithms yield partial and inaccurate explanation of the class distinction. Our algorithm gives a correct explanation, providing the noise part as the reason for classification. 
}
\label{fig:celeba_noise_xai_compare}
\end{figure}

In Figs. \ref{fig:qualitative_examples}, \ref{fig:celeba_noise_xai_compare} and \ref{fig:peppers} 
we show qualitative examples for various datasets. In general, our method is especially effective in cases where the differences between the classes have an additive nature. This occurs in situations where the differences involve textures (head- and facial-hair in \cref{fig:celeba_distinction_agnostic}), colors (\cref{fig:peppers}), and the presence or absence of details, such as cars (\cref{fig:cars_distinction_agnostic}), tumors (\cref{fig:brats_distinction_agnostic}) red lips or heavy eyebrows  (\cref{fig:celeba_distinction_agnostic}).


\begin{figure}[htb]
\begin{subfigure}{0.45\textwidth}
\centering
\includegraphics[width=1\linewidth]{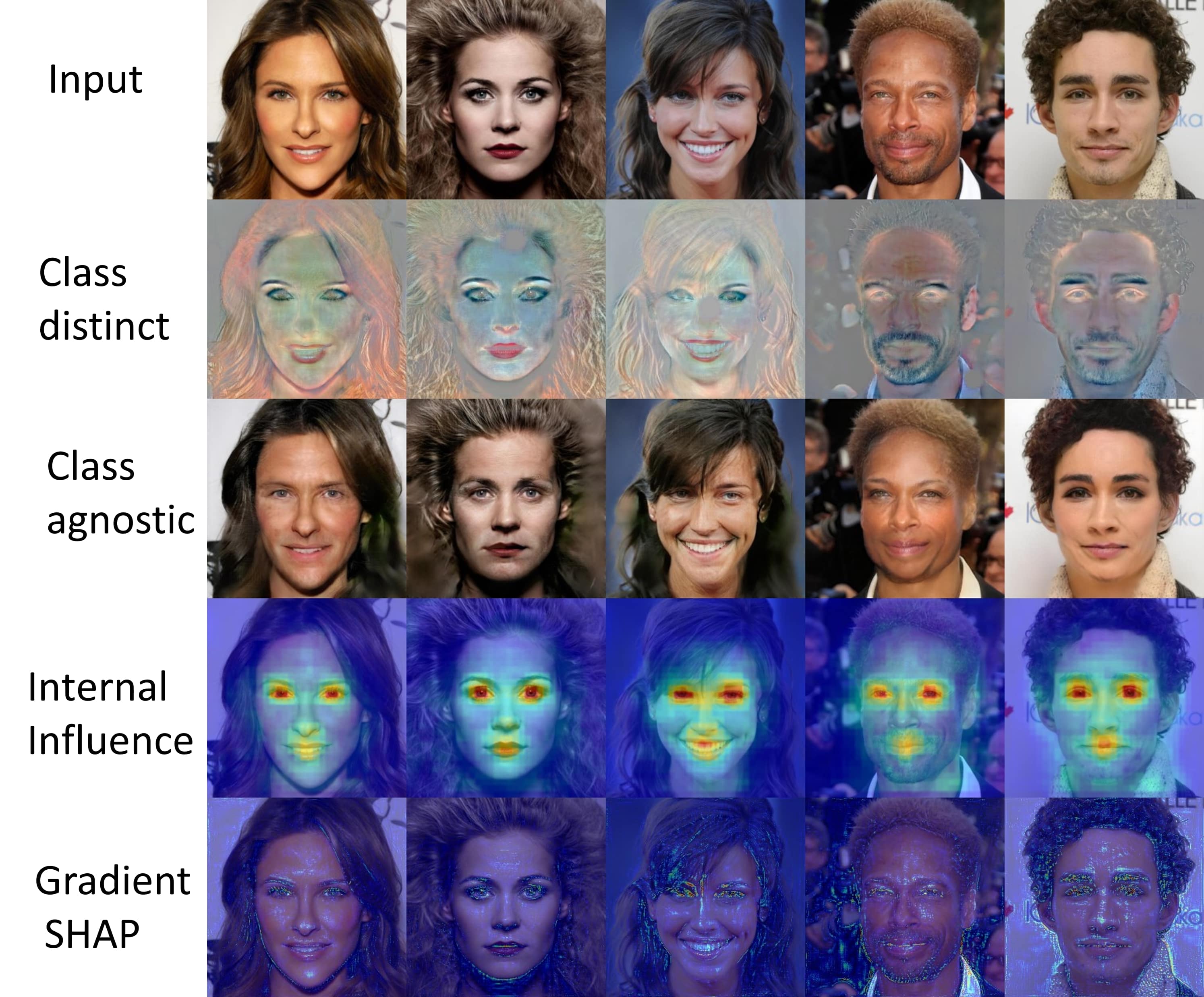}
\caption{CelebA \cite{guo2016ms}: The dataset included male and female classes.
}
\label{fig:celeba_distinction_agnostic}
\end{subfigure}
\begin{subfigure}{0.45\textwidth}
\centering
\includegraphics[width=1\linewidth]{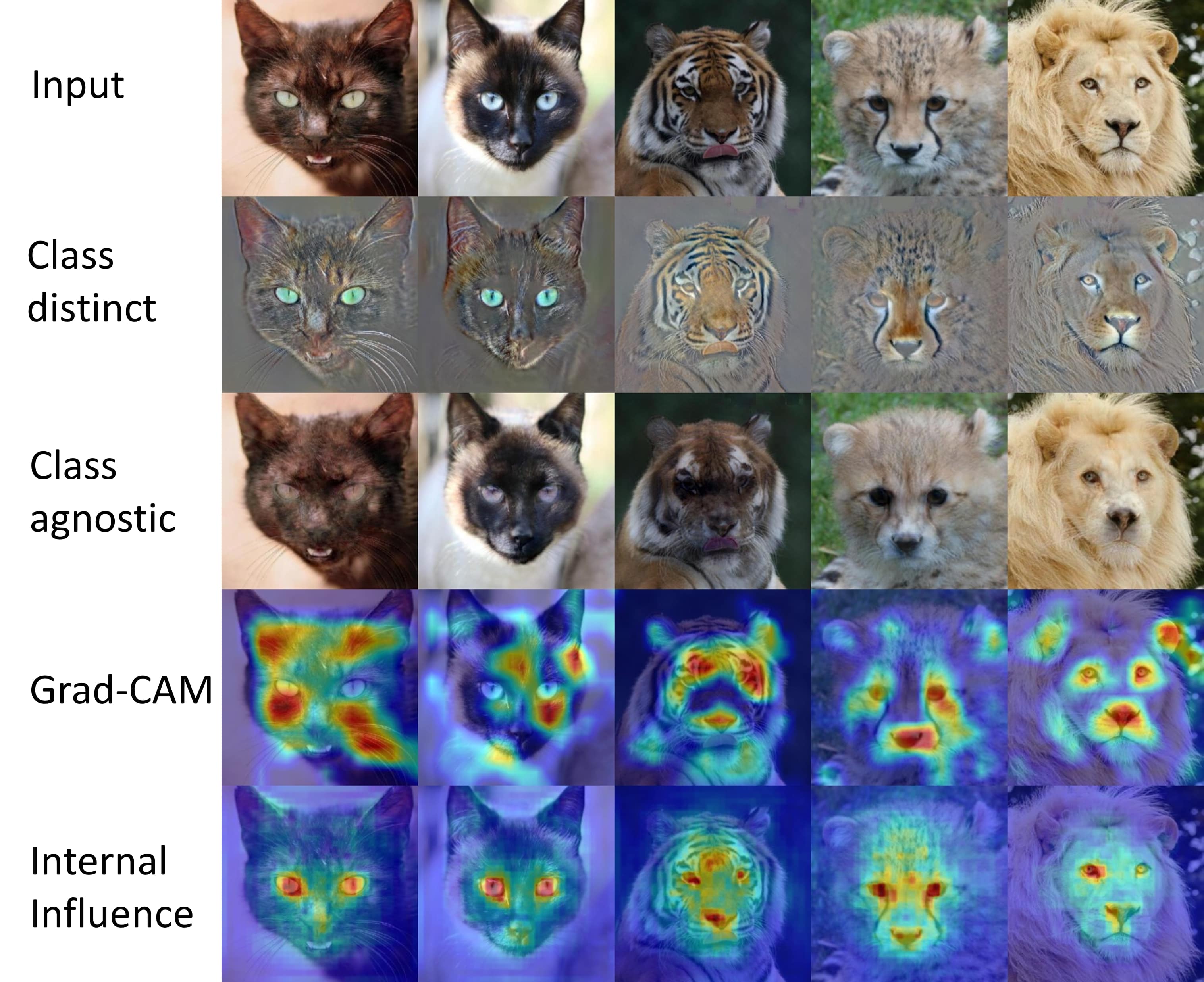}
\caption{AFHQ \cite{choi2020starganv2}: The dataset includes animals faces.
}
\label{fig:animals}
\end{subfigure}

\begin{subfigure}{0.45\textwidth}
\centering
    \includegraphics[width=1\linewidth]{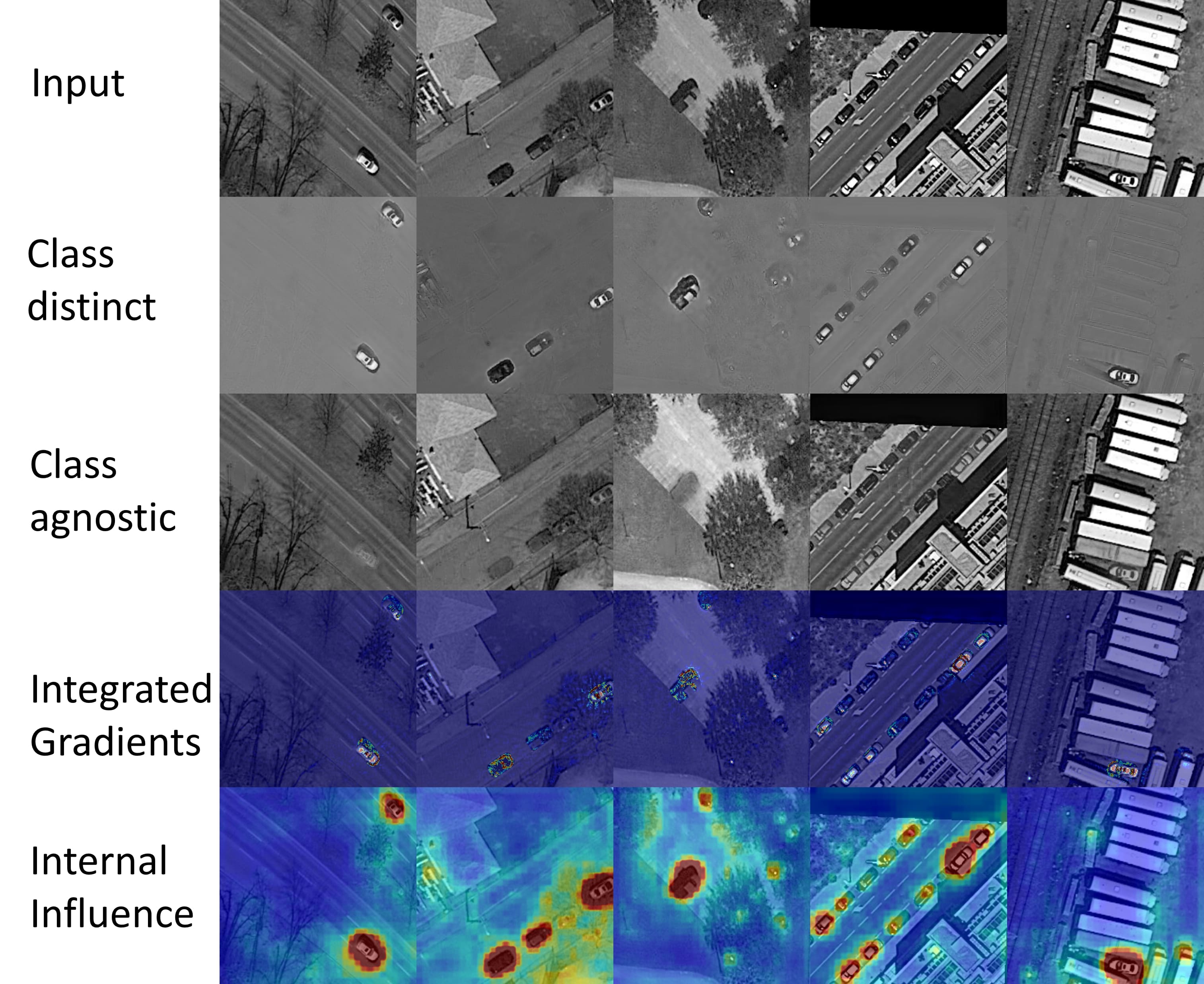}
    
    \caption{Cars (DOTA \cite{xia2018dota}): Aerial images. The dataset included two classes, images with and without cars.
    }
    \label{fig:cars_distinction_agnostic}
\end{subfigure}
\begin{subfigure}{0.45\textwidth}
\centering
    \includegraphics[width=1\linewidth]{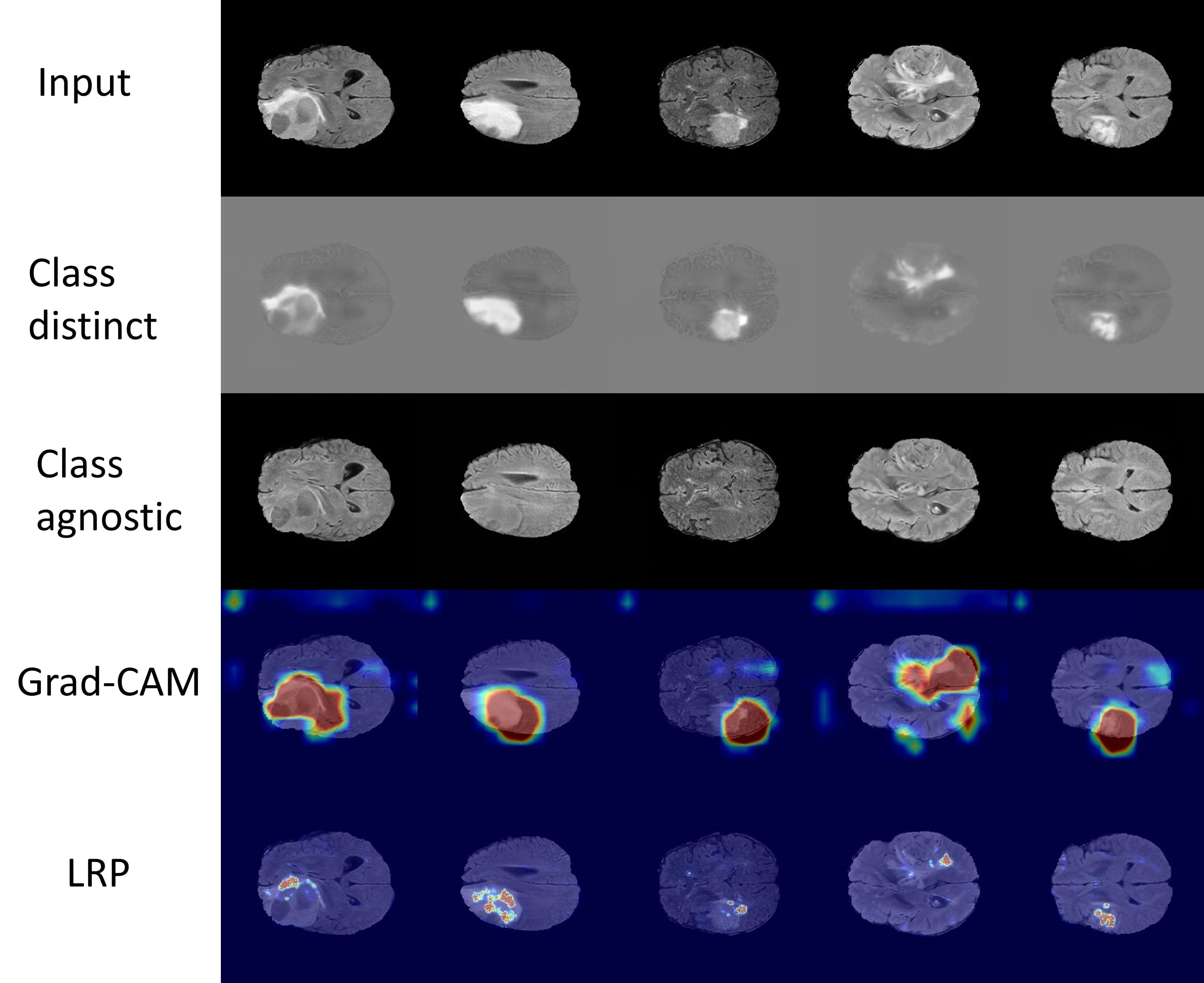}
    
    \caption{BraTS \cite{menze2014multimodal}: Tumors in MRI images. The dataset included two classes, with and without tumors.
    }
    \label{fig:brats_distinction_agnostic}
\end{subfigure}
\caption{Comparison of class-distinct and class-agnostic parts with heatmaps from other methods.}
\label{fig:qualitative_examples}
\end{figure}

\begin{figure}[htb]
\centering
\includegraphics[width=0.75\textwidth]{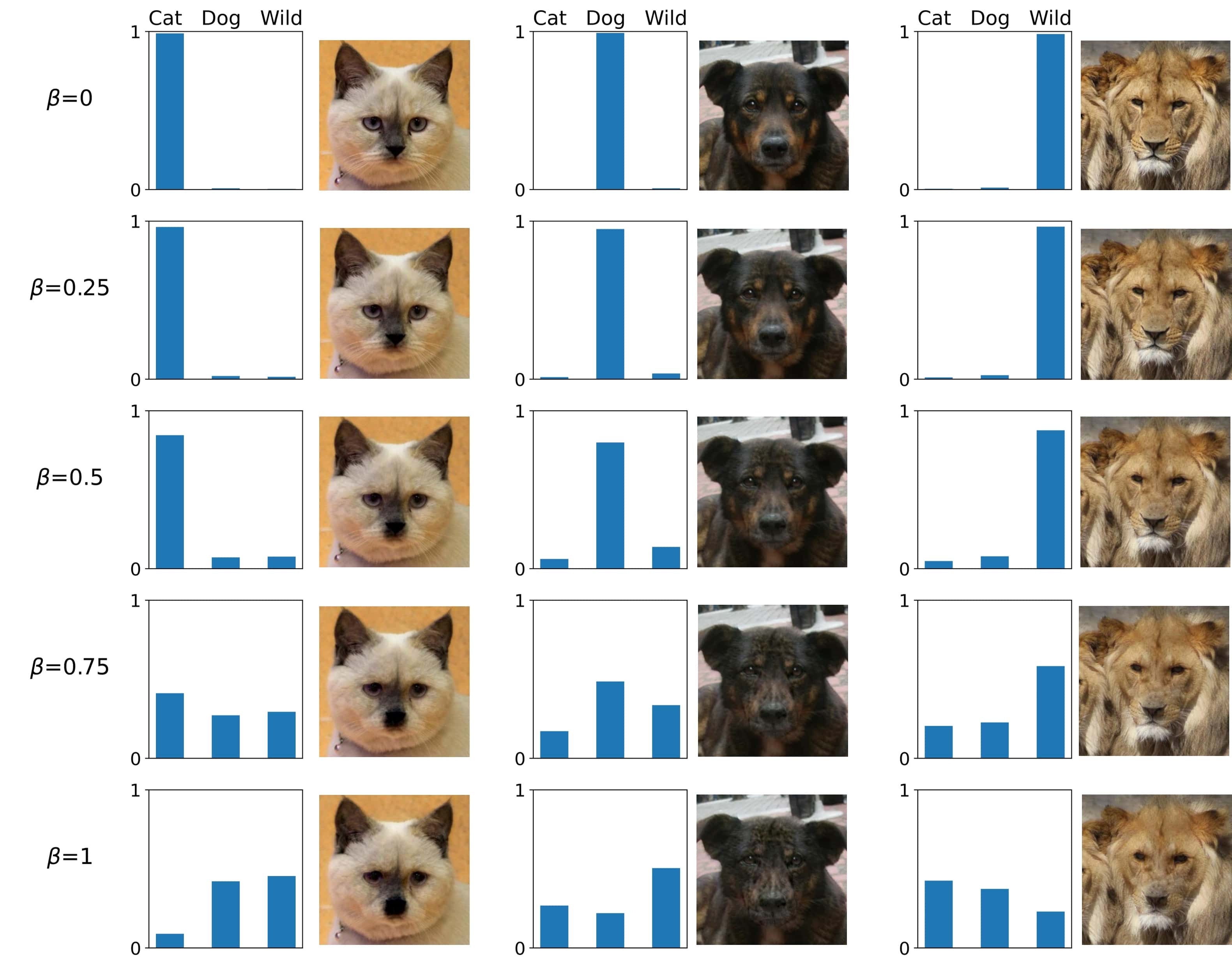}
\caption{{\bf Linear reduction in class distinction.} We examine the classification results of the following generated images $x_{\beta} = x - \beta \cdot \psi_{Distinct}$.
The bar plots show the average probabilities (softmax output) for each class, from left: Cat, Dog, Wild. An image example of $x_\beta$ is shown for each class. We observe that as $\beta$ increases, class distinction decreases, where for $\beta=1$ we have $x_\beta = \psi_{Agnostic}$. Ideally we should obtain equal probabilities for all classes, we observe our algorithm produces some overshoots toward alternative classes.
}
\label{fig:lin_class_reduction}
\end{figure}

In \cref{fig:lin_class_reduction} we show a linear progression from the original image to the agnostic part by generating the images,
\begin{equation}
    x_{\beta} = x - \beta \cdot \psi_{Distinct},
    \label{eq:x_beta}
\end{equation}
where $\beta \in \{0,0.25,0.5,0.75,1\}$. For $\beta=0$ we have the original image and for $\beta=1$ the agnostic part. The average probability vector $p(x_\beta)$ is depicted, averaged for each class. A single image for each class is shown, illustrating this progression. We see class distinction diminishes. Note we do not obtain precise agnostic images for $\beta=1$ but only an approximation.


\begin{table}[htb]
    \centering
    \caption{AUC of the curve: classification accuracy of $x_{\beta}$ as a function of $\beta$, see   \cref{eq:x_beta}, with $\beta$ ranging from 0 to 1. \emph{Lower AUC is better}. See full explanation in  \cref{sec:exp}.}
    
    \begin{tabular}[width=0.4\textwidth]{|l| c c c c c|}
    \hline
        ~ & \multicolumn{5}{c}{Datasets} \vline\\ \hline
        Method & Peppers & AFHQ & CelebA & BraTS & DOTA 
        \\ \hline
        Ours                 & \textbf{0.542} & \textbf{0.754} & \textbf{0.651} & \textbf{0.649} & \textbf{0.608} \\ 
        LRP \cite{binder2016layer}                 & 0.971 & 0.999 & 0.999 & 0.997 & 0.963 \\ 
        IG \cite{sundararajan2017axiomatic} & 0.881 & 0.908 & 0.946 & 0.934 & 0.828 \\
        G-SH \cite{kokhlikyan2020captum}       & 0.87  & 0.907 & 0.947 & 0.935 & 0.831 \\
        I-Inf. \cite{leino2018influence}    & 0.865 & 0.995 & 0.996 & 0.928 & 0.844 \\
        GGC \cite{selvaraju2017grad}      & 1.0   & 1.0   & 1.0   & 0.999 & 0.981 \\
        GC \cite{selvaraju2017grad}             & 0.966 & 1.0   & 0.999 & 0.996 & 0.901 \\
        Random               & 0.914 & 0.993 & 0.891 & 0.891 & 0.881 \\ \hline
    \end{tabular}
    \label{tab:global_AUC_compare}
\end{table}

In Table \ref{tab:global_AUC_compare} we show the results of a quantitative experiment comparing our method to other possible XAI decompositions. In this experiment we follow \cref{eq:x_beta} for  $\beta$ ranging from $0$ to $1$ in increments of $0.1$. As $\beta$ grows accuracy should drop. We check the area under the curve (AUC) of accuracy vs. $\beta$, where lower AUC is better. More details appear in the supplementary. 
Since, as far as we know, we are the first to propose DXAI, we obtain a decomposition based on established XAI algorithms, implemented using Captum library \cite{kokhlikyan2020captum}.  
One can produce a decomposition from a heatmap $H \ge 0$ by a normalization, having a weight for each pixel,
$ w = \frac{H}{\max (H)} \in [0,1],$
and defining for an image $x$, $ \psi_{Distinct} := w \cdot x$, $\psi_{Agnostic} := (1-w) \cdot x$. We get a deomposition in the form of \cref{eq:xai_decomp},
such that $x = \psi_{Distinct} + \psi_{Agnostic}$, see examples in \cref{fig:xai_compare}. Table \ref{tab:global_AUC_compare} demonstrates that for all data sets our decomposition outperforms all other methods by a considerable margin. This indicates our proposed method has evidently different qualitative properties, such that trivial manipulations of the heatmap cannot generate high quality class-distinct components.

Our algorithm does not provide importance ranking of each pixels, with respect to its contribution for the classification. This is one of its limitations. Thus we cannot use the standard way to evaluate XAI accuracy by removing pixels gradually based on importance, as done e.g. in \cite{samek2016evaluating}.
In some problems, however, a simple ranking can be inferred by our algorithm. In a binary classification problem, when one class is predominantly decided based on the existence of certain lighter pixels, we can use the amplitude of pixels in $\psi_{Distinct}$ as a reasonable importance ranking. In this case standard XAI evaluation can be made. The BraTS dataset \cite{menze2014multimodal} is such a case. It contains MRI scans of the human brain used for brain tumor segmentation research. The dataset includes scans from patients with brain tumors, along with expert annotations for tumor regions. In many cases, bright regions indicate evidence of tumors. We divided the dataset into two classes: images containing tumors and images that do not. Some example results are shown in  \cref{fig:brats_distinction_agnostic}. 
 \Cref{tab:BraTS} shows the standard AUC evaluation on this set (AUC of accuracy vs. number of pixels removed, ordered by importance). We see our algorithm behaves favorably both qualitatively and quantitatively. 

\begin{table*}[!h]
    \centering
    \caption{AUC comparison of the Faithfulness index on the BraTS dataset among various XAI methods. 
    Ours is compared to LRP, Integrated gradients, Gradient SHAP, Internal influence, Guided Grad-Cam, Grad-Cam and random noise. AUC values were calculated based on the deletion of pixels, with up to 20\% removed.
    When our CD part may be interpreted as importance (by pixel amplitude) one can validate our accurate XAI mapping also by standard evaluation methods.
    }
    \begin{tabular}{|l c c c c c c c|}
    \hline 
        Ours & LRP \cite{binder2016layer} & IG \cite{sundararajan2017axiomatic} & G-SH \cite{kokhlikyan2020captum} & I-Inf \cite{leino2018influence} & GGC \cite{selvaraju2017grad} & GC \cite{selvaraju2017grad} & RAND 
        \\ \hline
        \textbf{0.125} &0.156 & 0.171 & 0.172 & 0.177 & 0.179 & 0.192 & 0.195
        \\ \hline
    \end{tabular}
    
    \label{tab:BraTS}
\end{table*}

\begin{table*}[!h]
    \centering
    \caption[Matching explanation to classifier.]{{\bf Matching explanation to classifier.} At each row DXAI training is performed using a different classifier. AUC (similiar to \cref{tab:global_AUC_compare}) is shown, computed using the two classifiers, on four different data sets. AUC is smaller (better) when the classifier in training and the one used to compute the AUC match.
    This shows the relevance of the DXAI explanations to the classifier.}
    \begin{small}
    \begin{tabular}[width=0.9\textwidth]{|c|c c | c c|}
    \hline
        ~ & \multicolumn{2}{c}{BraTS} \vline & \multicolumn{2}{c}{DOTA (cars)} \vline \\ \hline
        AUC according to: & ResNet18 & Simple & ResNet18 & Simple \\  
        Trained by ResNet18          & \textbf{0.49} & 0.787 & \textbf{0.793} & 0.903 \\ 
        Trained by Simple-Classifier & 0.629 & \textbf{0.562} & 0.914 & \textbf{0.729} \\ \hline

        ~ & \multicolumn{2}{c}{AFHQ} \vline & \multicolumn{2}{c}{Celeb A} \vline \\ \hline
        AUC according to: & ResNet18 & Simple & ResNet18 & Simple \\  
        Trained by ResNet18          & \textbf{0.61} & 0.871 & \textbf{0.606} & 0.807 \\ 
        Trained by Simple-Classifier & 0.917 & \textbf{0.546} & 0.809 & \textbf{0.591} \\ \hline
    \end{tabular}
    \end{small}
    \label{tab:compare_classifiers}
\end{table*}

An XAI algorithm should naturally depend on the specific classifier $C$ at hand. Different classifiers may yield different class distinct and class agnostic parts. We check this is indeed the case in Table \ref{tab:compare_classifiers}.
We compare our results when $C$ is 
ResNet18 
and when $C$ is a simpler classifier with fewer layers, referred to as ``Simple'' (yielding less accurate results, details in the supplementary). 
We compare AUC as done for \cref{tab:global_AUC_compare}. Here however, the accuracy graph is computed twice - using each of the above classifiers. We show AUC drops more sharply when $C$ used for obtaining $\psi_{Distinct}$ in \cref{eq:x_beta} matches the
classifier for computing the AUC.
The experiment demonstrates that our map effectively captures the characteristics of the target classifier.

\begin{table}[htb]
    \centering
    \caption{\textbf{Stability experiments.} We tested the algorithm on several datasets with varied initializations and measured the standard deviation (STD) between the distinct maps. In the left table, we show a relatively small variation between solutions despite using a generative model, indicating consistent convergence. In the right table, we compared the normalized STD by dynamic range to two other XAI methods. Our approach calculates the STD over three different initializations, while in the others over three sequential layers. Notably, the layer of the classifier impacts the explanations, serving as a hyperparameter. Our method demonstrates relative stability, detailed further in the supplementary materials.
    }
    \begin{tabular}[width=0.4\textwidth]{|c|c c c c|}
    \hline
    \multicolumn{5}{|c|}{STD during Iterations} \\ \hline
    Iteration & 1 & 10K & 100K & 300K\\ \hline
    Celeb A & 0.33 & 0.094 & 0.032 & 0.027\\
    AFHQ & 0.332 & 0.101 & 0.082 & 0.066\\
    Peppers & 0.312 & 0.099 & 0.035 & 0.052\\ \hline
    \end{tabular}
    \begin{tabular}[width=0.4\textwidth]{|c| c c c|}
    \hline
    \multicolumn{4}{|c|}{Normalized STD by Range} \\ \hline
    Methods: & Ours & GC \cite{selvaraju2017grad} & I-Inf \cite{leino2018influence} \\ \hline
    Celeb & \textbf{0.024} & 0.098 & 0.072\\
    AFHQ & \textbf{0.05} & 0.126 & 0.062 \\
    Peppers & \textbf{0.041} & 0.223 & 0.21 \\ \hline
    \end{tabular}
    
    \label{tab:stability}
\end{table}

\section{Discussion and Conclusion}
\label{sec:conclusion}

We propose an alternative way to analyze and to visualize the reasons for classification by neural networks. It is based on decomposing the image into a part which does not contribute to the classification and to one which holds the class-related cues. 
This approach may not be ideal for all applications and has several \emph{limitations and drawbacks}, compared to standard XAI methods: 
    1. The method requires training, for a specific training set and classifier (training is slow, inference is fast); 
    2. There is no natural ranking of the significance of pixels in the image. The amplitude of pixels in the class-distinct part can serve as a good approximation; 
    3. Our implementation uses GANs. 
    The proposed concept does not rely on a GAN architecture and improvements may be achieved by diffusion-type  generative models, such as \cite{jeanneret2022diffusion, yang2023zero, wang2023stylediffusion, chen2023controlstyle}.

Major advantages afforded by this method are of  obtaining detailed, dense, high-resolution, multi-channel information for class distinction. We propose, for the first time, a decomposition problem for XAI purposes, allowing to clearly visualize also image components which are not relevant for the classification task (agnostic part). We examine and compare our method in the context of several classification tasks and data sets, showing the applicability and additional insights afforded by this new approach.

%
%
\clearpage
\bibliographystyle{splncs04}
\bibliography{main}

\clearpage
\appendix
\newcommand{\supp}{{\huge \textbf{Appendix}} \xspace}
\supp 
\section{Datasets}
\label{datasets}
In the main paper several datasets were used. We provide some more technical details on those sets. In \cref{tab:datasets_size} the number of images of each set is given. Some datasets were adjusted or filtered to suit the classification task. For instance, in the case of DOTA \cite{xia2018dota} , which contains high-resolution aerial images with objects marked in bounding boxes, we divided the images into patches of size 256x256 pixels and categorized them into two classes. One class comprised images featuring at least 70\% of a single car (in many cases there are multiple cars), while the other class included images with no cars at all. The classifier was trained to distinguish between these two classes of patches. As expected, the explanation (distinct) image is comprised of isolated cars. 

\begin{table}[htb]
    \centering
    \begin{tabular}{|l|c c|}
    \hline
        Datasets & Train set & Test set \\ \hline
        AFHQ \cite{choi2020starganv2} & 14630 & 1500 \\
        CelebA \cite{guo2016ms} & 28000 & 2000 \\ 
        DOTA(cars) \cite{xia2018dota} & 172465 & 20696 \\
        BraTS \cite{menze2014multimodal} & 15000 & 1232 \\
        Peppers \cite{thi2019fruits} & 2478 & 826 \\
        Tomatoes \cite{thi2019fruits} & 5103 & 1707 \\
        Apples \cite{thi2019fruits} & 6404 & 2134 \\ \hline
    \end{tabular}    
    \caption{Number of images in each dataset.}
    \label{tab:datasets_size}
\end{table}

To identify fruits, we employed the Fruits 360 dataset \cite{thi2019fruits}. From there, we selected specific classes of peppers, tomatoes, and apples, which mainly differ in color and texture, and created a dataset for each.

For BraTS \cite{menze2014multimodal}, we only considered a subset of the images—specifically, those in which the tumor occupied an area of at least 20 pixels were included in the class designated to contain tumors. Images that did not contain a tumor at all but featured a full cross-section of the brain were included in the second class. Subsequently, we randomly split the images into train and test sets.

No modifications were made to the remaining datasets.

\section{Classifiers}
\label{classifiers}
In our experiments three different classifiers were employed for the classification block $C$: a ResNet18-type classifier, a  simple classifier described in \cref{tab:simple-classifier}, and our discriminator $D$, which possesses classification capabilities, as the output is a vector of length $c$, indicating class probabilities (details of the architecture are in \cref{tab:discriminator architecture}). The test accuracy of each classifier is given in \cref{tab:classifiers}. We note that the discriminator exhibits the best classification performance. Thus, we selected it in the main paper to show the explanation images by the various DXAI and XAI algorithms. 
We conducted experiments also on ResNet18 and on the Simple classifier, showing consistently similar trends. We provide later some examples of ResNet18 results on the same datasets.

\begin{table}[htb]
    \centering
    \begin{tabular}{|l|c|c|c|}
    \hline
        \textbf{Datasets} & \textbf{ResNet18} & 
        \textbf{Simple} & 
        \textbf{Discriminator} 
        \\ \hline
        AFHQ & 0.975 & 0.906 & 0.994 \\
        CelebA & 0.978 & 0.96 & 0.988 \\
        DOTA(cars) & 0.863 & 0.749 & 0.858 \\
        BraTS & 0.912 & 0.877 & 0.982 \\
        Peppers & 1 & 0.944 & 1 \\ 
        Tomatoes & 1 & 0.985 & 1 \\
        Apples & 0.983 & 0.909 & 1 \\ \hline
    \end{tabular}
    \caption{Test accuracy of each classifier.}
    \label{tab:classifiers}
\end{table}

\begin{table}[htb]
    \centering
    \begin{tabular}{|c|c|}
    \hline
        \textbf{Simple-Classifier Architecture} & \textbf{Dimension} \\ \hline
        Input image & 256x256x3 \\
        Conv2d(3, 16, 3x3) & 256x256x16 \\
        MaxPool2d(3x3) & 85x85x16 \\
        ReLU & ~ \\
        Conv2d(16, 32, 3x3) & 85x85x32 \\
        MaxPool2d(3x3) & 28x28x32 \\
        Linear(in=28*28*32, out=128) & 128 \\
        ReLU & ~ \\
        Linear(in=128, out=c) & $c$ \\ \hline
    \end{tabular}
    \caption{Simple-Classifier Architecture from input to output. $c$ is the number of classes and input image can be also a gray scale image.}
    \label{tab:simple-classifier}
\end{table}

\begin{table}[htb]
    \centering
    \begin{tabular}{|c|c|}
    \hline
        \textbf{Disriminator Architecture} & \textbf{Dimension} \\ \hline
        Input image & 256x256x3 \\
        Conv1x1 & 256x256x64 \\
        ResBlk & 128x128x128 \\
        ResBlk & 64x64x256 \\
        ResBlk & 32x32x512 \\
        ResBlk & 16x16x512 \\
        ResBlk & 8x8x512 \\
        ResBlk & 4x4x512 \\
        Leaky-ReLU & ~ \\
        Conv4x4 & 1x1x512 \\
        Leaky-ReLU & ~ \\
        Conv1x1 & $c$ \\ \hline
    \end{tabular}
    \caption{Discriminator Architecture as given in \cite{choi2020starganv2} from input to output. $c$ is the number of classes and unlike the original architecture the input image can be also a gray scale image.}
    \label{tab:discriminator architecture}
\end{table}

The intentionally designed simple classifier, with lower classification ability and simple architecture, provides distinct classification results from a well-known classifier like ResNet18. This intentional dissimilarity allows for meaningful comparisons of their results and verify the connection between the classifier and its explanation, as detailed in the main paper.

\section{Training - Additional Details}
\label{training additional details}
In the main paper, we presented the training method and primary loss functions. However, for brevity, we omitted details on functions that are neither innovative nor crucial for understanding the method's general concept. Additionally, these functions serve similar purposes to those already discussed.

One such function, integral to the GAN architecture implementation but not explicitly mentioned, is the adversarial loss as defined in \cite{choi2020starganv2} 

\begin{equation}
    L_{adv} = \mathop{\mathbb{E}}_{x,y}\left[log D_y(x^y)\right] + \mathop{\mathbb{E}}_{x,\tilde{y}}\left[log(1-D_{\tilde{y}}(\hat{x}^{\tilde{y}}))\right],
    \label{eq:adv_loss}
\end{equation}
where $D_y$ represents the $y$'th element of a vector of length $c$, which is an output of the discriminator. Its role is to ensure that the generators produce images resembling real ones, for a given class. 

Another function, briefly mentioned, directs our discriminator not only to classify images but also to align with the classifications of our trained classifier. This encourages the discriminator to assign a high score only to the correct class according to the classifier  
\begin{equation}
    L_{KLD} =KLD\Bigl(D(x^y), C(x^y) \Bigl),
    \label{eq:KLD}
\end{equation}
where $KLD$ is the Kullback-Leibler distance.

In addition to the reconstruction loss explained in the main part, one can use additional constraints on the reconstruction to enhance results. Specifically, we observed challenges in reproducing areas with significant differences between classes. To address this, we incorporated an additional constraint for reconstruction between pixels with a high amplitude in the distinction branch. High amplitude signifies differences between classes due to the additive nature of the model. We remind that we use the following distance measure 
\[
    d(u,v) = \|u-v\|_{L^1}+\|u-v\|_{L^2},
\] 
with a reconstruction loss of the form
\[
    L_{rec} = d(x,\hat{x}^y) + d(x,\hat{x}^{\tilde{y}}).
\]
The proposed loss function is 
\begin{equation}
    L_{dis-rec} = d(x \odot \mathbb{I},\hat{x}^y \odot \mathbb{I}),
    \label{eq:dis-rec}
\end{equation}
where
\[
    \mathbb{I} = 
        \begin{cases} 
            1 & \text{if $|\psi_1^{y}|>mean(|\psi_1^{y}|)$} \\
            0  & \text{else},
        \end{cases}
\]
and $\odot$ denotes element-wise product.

The full objective loss is:
\begin{equation}
\begin{gathered}
    L_{Total} = 
     \lambda_{adv}L_{adv} + \lambda_{KLD}L_{KLD} +\lambda_{cf}L_{class-fake} \\
     + \lambda_{rec}L_{rec} + \lambda_{dr}L_{dis-rec}
 \end{gathered}
\end{equation}
In our experiments we used the following weights:
\begin{equation}
\begin{gathered}
     \lambda_{adv}=2,  \lambda_{KLD}=1,  \lambda_{cf}=2, \\
     \lambda_{rec}=4,  \lambda_{dr}=4.
 \end{gathered}
 \end{equation}
 
A note on how weights can be adjusted.
The rule of thumb is that for applications where particularly good reconstruction is required, the weights for reconstruction should be increased. Conversely, when style transition is challenging and there are more hidden characteristics, the weights for classification and adversarial loss should be increased.

\begin{figure}[htb]
    \centering
    \includegraphics[width=0.8\textwidth]{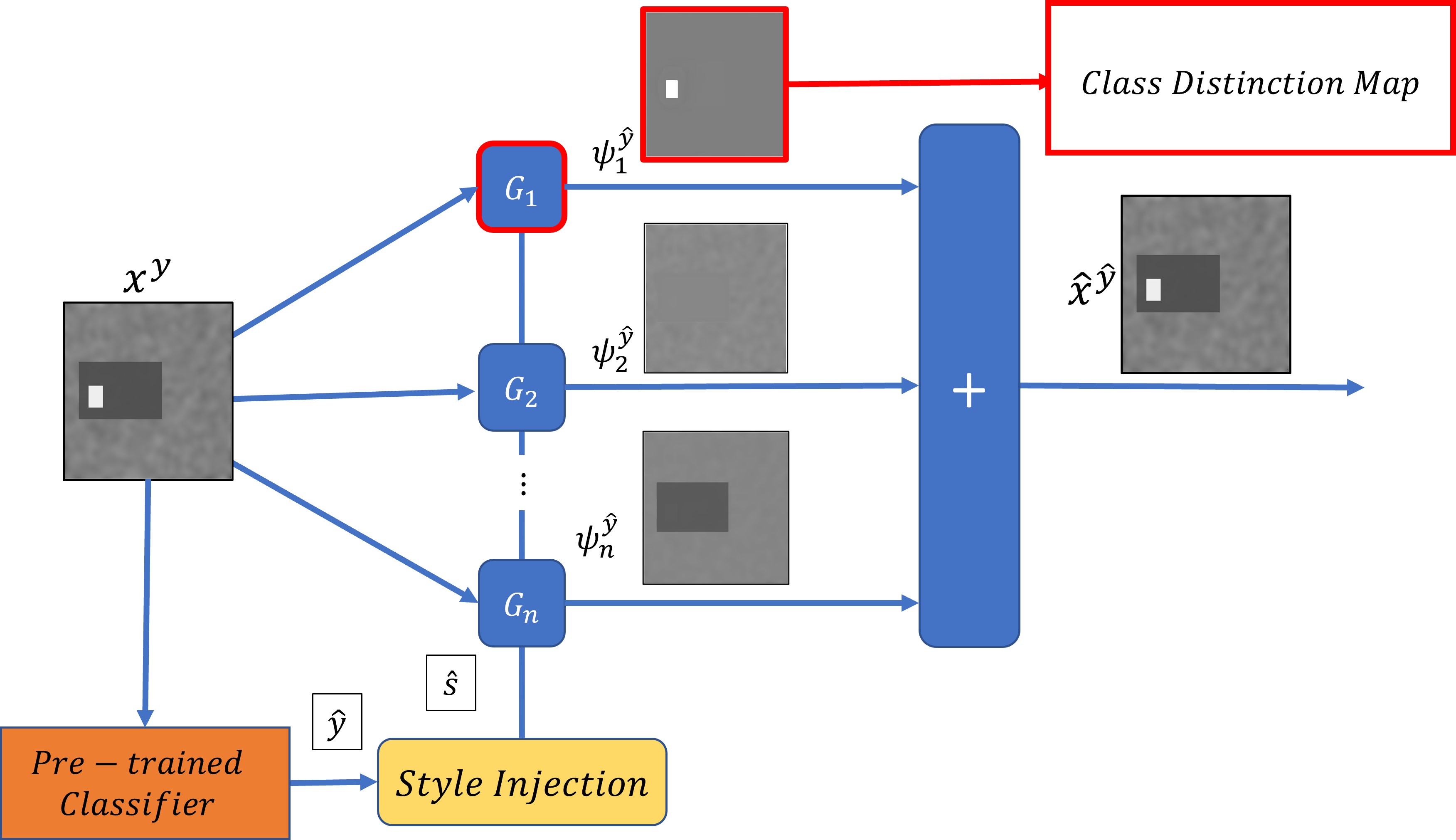}
    
    \caption{Inference stage: the input image and the predicted class are used to guide the generators. The output is comprised of the class distinction map in the first branch and the agnostic part by a sum of  the rest of the branches.}
    \label{fig:inference_stage}
\end{figure}

\section{Inference Stage}
\label{inference}
The inference stage is described in \cref{fig:inference_stage}. The input $x^y$ is an image $x$ of class $y$. The class predicted by the classifier is $\hat{y}$. We would like to explain the classification result by the trained DXAI model. The class distinct 
part is given by the first generator. The rest of the summed branches provide the agnostic part.


\section{Ablation}
\label{ablation}

\begin{figure}[htb]
    \centering
    \includegraphics[width=0.9\textwidth]{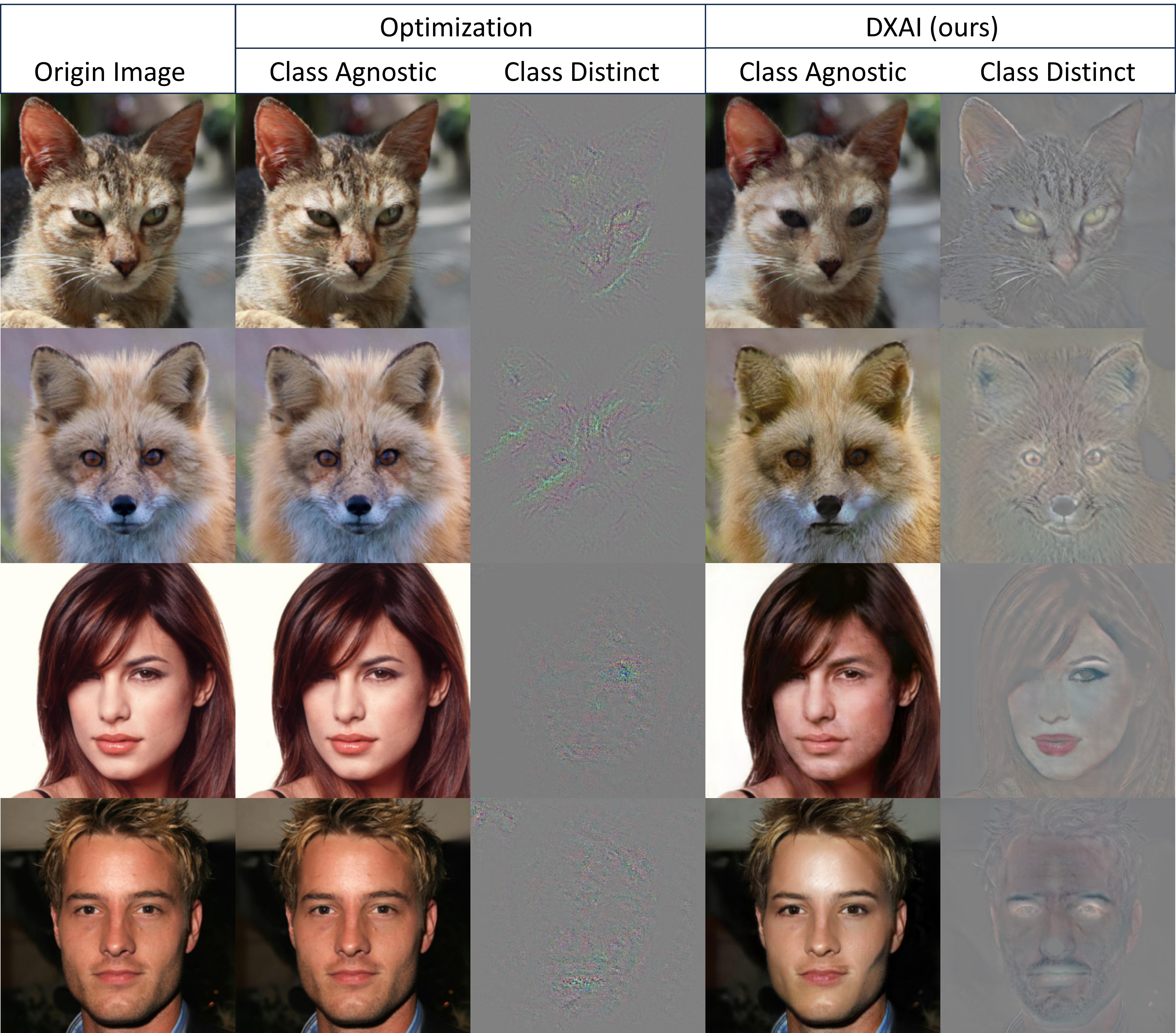}
    \caption{Agnostic and distinct components in two cases. Middle columns, the agnostic part is found through optimization. In this case, the distinct part resembles more noise and has little semantic features . Right columns, our method which provides much more structured and semantic components, containing features that represent the predicted class.}
    \label{fig:optimization_vs_dxa}
\end{figure}

In the ablation study, we tried the naive method to obtain the class distinct component. As shortly mentioned in the main paper, apparently it's possible to obtain it by optimization. The idea is to get the agnostic component and subtract it from the original image. The optimization is done as follows: Initiate $\psi_{Agnostic}=x$ when $x$ is an image from our data.
Then, get the predicted probability of each class by entering the image into the classifier:
$$p = C(\psi_{Agnostic}).$$
Next, calculate the KL distance between the obtained distribution and the uniform distribution as follows:
$$
D_{KL} = \sum\limits_{i=0}^{c-1} \frac{1}{c}(\log \frac{1}{c} - \log p_i).
$$
Finally, compute iteratively or until convergence:
$$
\psi_{Agnostic} = \psi_{Agnostic} - dt \nabla_{\psi_{Agnostic}} D_{KL},
$$
and the distinct component will be $\psi_{Distinct}=x- \psi_{Agnostic}$. 
We received that the process does converge quickly to images for which the output is approximately a uniform distribution (With $D_{KL}<10^-6$). However, the results are not very informative, as can be seen in \cref{fig:optimization_vs_dxa}.
Note that this is a highly non-convex problem where many local minima are possible.
In the optimization case we get a result which is with very little semantic meaning. It is closer, in some sense, to adversarial attacks.
This is in contrast to our method, where the results are based on the entire training set. The generative process learns how to produce features that, on average, will confuse the classifier and therefore are semantic in nature. We believe that this distinction may lead to fruitful future research, also in the case of robust network analysis and defense.

In addition, we investigated two more key factors. Firstly, we examined the impact of the number of branches. While decomposing the classified image into several images, we primarily focus on solving only two images—distinct and agnostic. One might question why not use only two branches for the solution, since two branches can be trained to achieve a similar solution.

\begin{figure}[htb]%
    \begin{subfigure}{0.45\textwidth}
    \centering
    \includegraphics[width=1\textwidth]{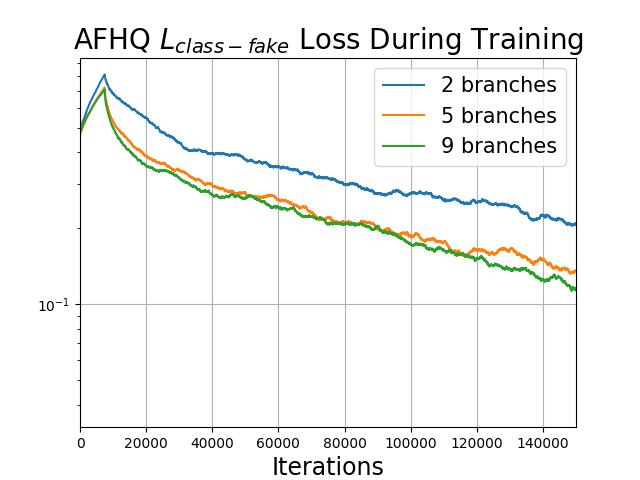}
    \caption{AFHQ - Class-fake loss}
    \end{subfigure}
    \begin{subfigure}{0.45\textwidth}
    \centering
    \includegraphics[width=1\textwidth]{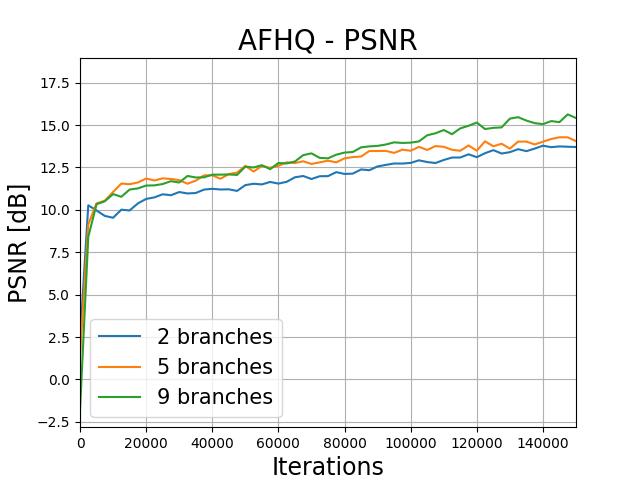}
    \caption{AFHQ - PSNR}
    \end{subfigure}

    \begin{subfigure}{0.45\textwidth}
    \centering
    \includegraphics[width=1\textwidth]{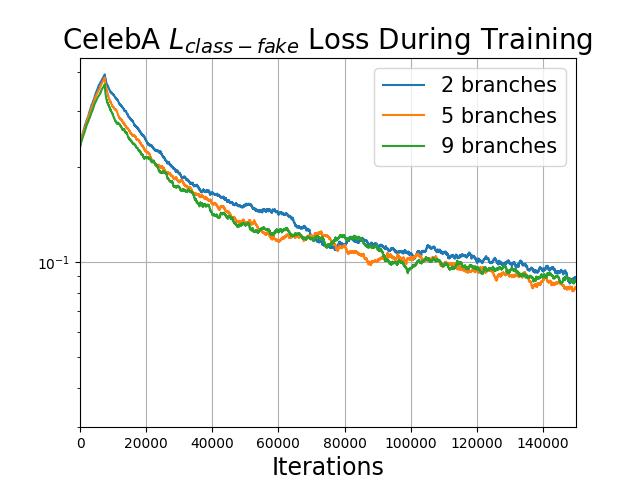}
    \caption{CelebA - Class-fake loss}
    \end{subfigure}
    \begin{subfigure}{0.45\textwidth}
    \centering
    \includegraphics[width=1\textwidth]{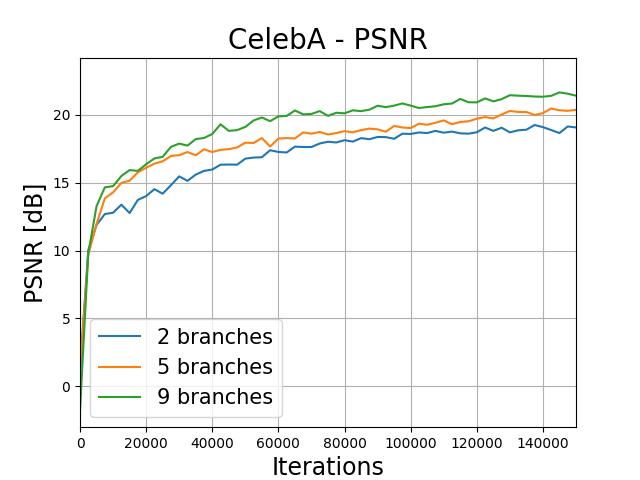}
    \caption{CelebA - PSNR}
    \end{subfigure}
    \caption{AFHQ \& CelebA ablation - need of more than 2 branches. The run of the experiment was stopped after 150K iterations, after the tendency was understood.}%
    \label{fig:2_branch_ablation_128}%
\end{figure}

We show that when branches are used for the agnostic part results are better both in terms of reconstruction and of the generators' ability to produce images that explain the classifier. For instance, as demonstrated in \cref{fig:2_branch_ablation_128}, PSNR decreases when using only two branches. Additionally, the loss $L_{class-fake}$, representing the generators' ability to produce meaningful images of a specific class according to the given classifier, is higher for the entire training with two branches. In other words, the classifier interprets the images less accurately as the desired class, making the images less reliable.
The additional generators provide better generation capacity and can be trained in an easier and more stable manner.

\begin{figure}%
    \centering
    \subfloat[\centering CelebA PSNR]{{\includegraphics[width=0.45\textwidth]{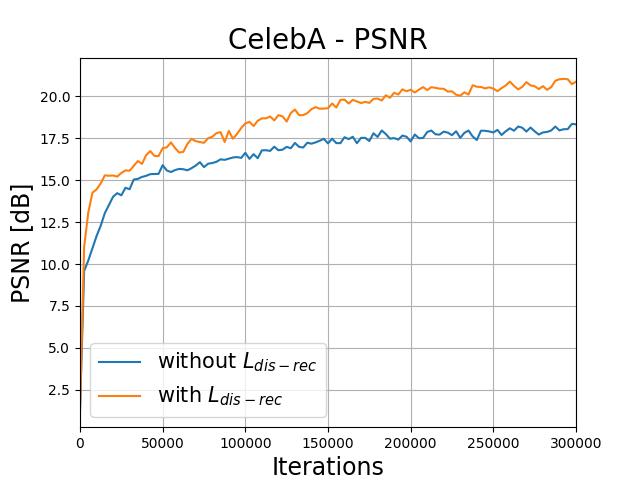} }}%
    \qquad
    \subfloat[\centering AFHQ PSNR]{{\includegraphics[width=0.45\textwidth]{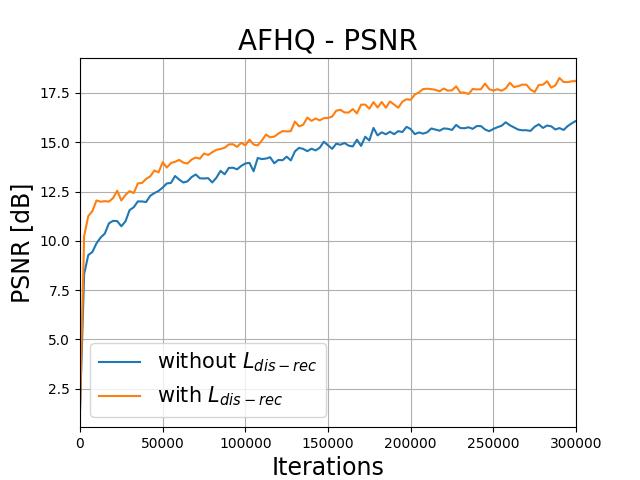} }}%
    \caption{Ablation of $L_{dis-rec}$, it can be seen that without this loss function the PSNR is reduced}%
    \label{fig:dis_rec_ablation}%
\end{figure}

In addition, we evaluated the impact of the loss $L_{dis-rec}$ described in \cref{eq:dis-rec} on the reconstruction quality. As explained earlier, we employed it because we observed that reconstruction, especially in areas with differences between the classes, was challenging. We conducted experiments both with and without it. We show that it indeed contributes to the quality of the reconstruction in terms of PSNR, as illustrated in \cref{fig:dis_rec_ablation}.

It should be noted that, due to time constraints, we did not run the ablation experiments until the convergence of an optimal solution. Instead, we stopped the training after observing a discernible trend in the change of solution quality.

\section{Stability Experiments Details}

As described in the main paper, we utilize image decomposition to address various issues encountered in existing XAI methods. To ensure adaptability across different classifiers and data types, we employ generative models for generating image decompositions. However, we are aware that training using the gradient descent method does not guarantee convergence, raising concerns about the consistency of solutions across different model initializations.

 \begin{figure}[htb]
    \centering
    \includegraphics[width=0.8\textwidth]{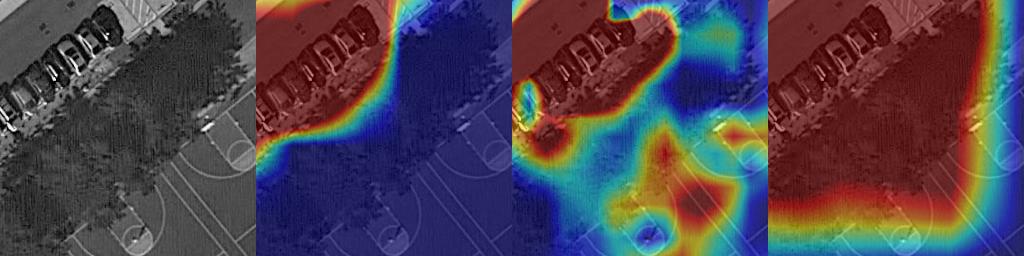}
    \caption{Instability observed in Grad-CAM example with sequential layers (from \cref{fig:Grad-CAM displaying} in the main paper).}
    \label{fig:grad_cam_unstable}
    \end{figure}

To mitigate these concerns, we conducted experiments (Table 4) to examine whether the explanations provided by the network for classifications (specifically, ResNet18 in our experiment) become increasingly consistent throughout training, as indicated by the standard deviation of the explanations for different initializations.

Additionally, we compared this standard deviation with that of two other algorithms (Grad-CAM\cite{selvaraju2017grad} and Internal Influence\cite{leino2018influence}), which also yield correct but potentially different solutions based on different choices of hyperparameters (in our case, the activation layer of the classifier we aim to explain). We conducted these experiments with three different initializations and three sequential layers to calculate the standard deviation.

Since the values of the pixels provided by the solutions vary between different methods, for example, in our method, the range of values is from minus 1 to 1, compared to the other two methods where the values are positive and theoretically between 0 and 1. Therefore, we normalized the standard deviation of each image within the dynamic range of the solutions, which is approximately the maximum value obtained from the three different solutions minus the minimum value. This normalization allows for a fair evaluation of quality.

Our experiments demonstrated that the standard deviation decreases with training, indicating convergence toward a consistent solution. Moreover, our standard deviation is relatively small compared to the alternatives.


\section{More results}

\begin{figure*}[htb]
    \centering
    \includegraphics[width=1\textwidth]{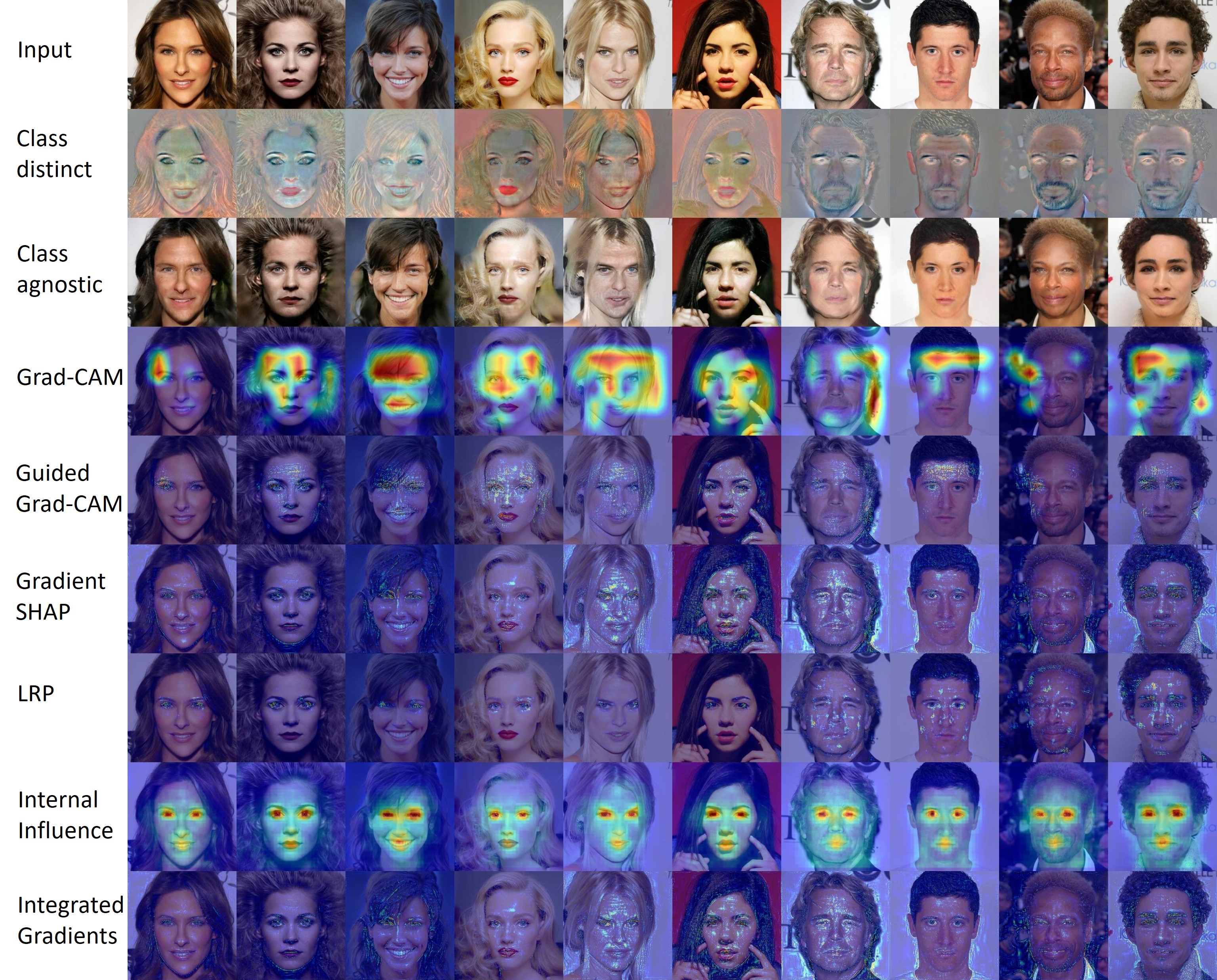}
    \caption{CelebA (Female and Male classes, $C=D$). The class distinct part shows for Female distinction by red lips, hair texture, eye-lashes and eyebrows. Fro Male -- facial hair, heavy eyebrows, strong chin and visibility of ears. Other XAI algorithms capture only parts of the full explanation for such a complex classification task, which involves many subtle cues.}
    \label{fig:celeb_all_heatmaps_compare_disc}
\end{figure*}

\begin{figure*}[p]
    \centering
    \includegraphics[width=1\textwidth]{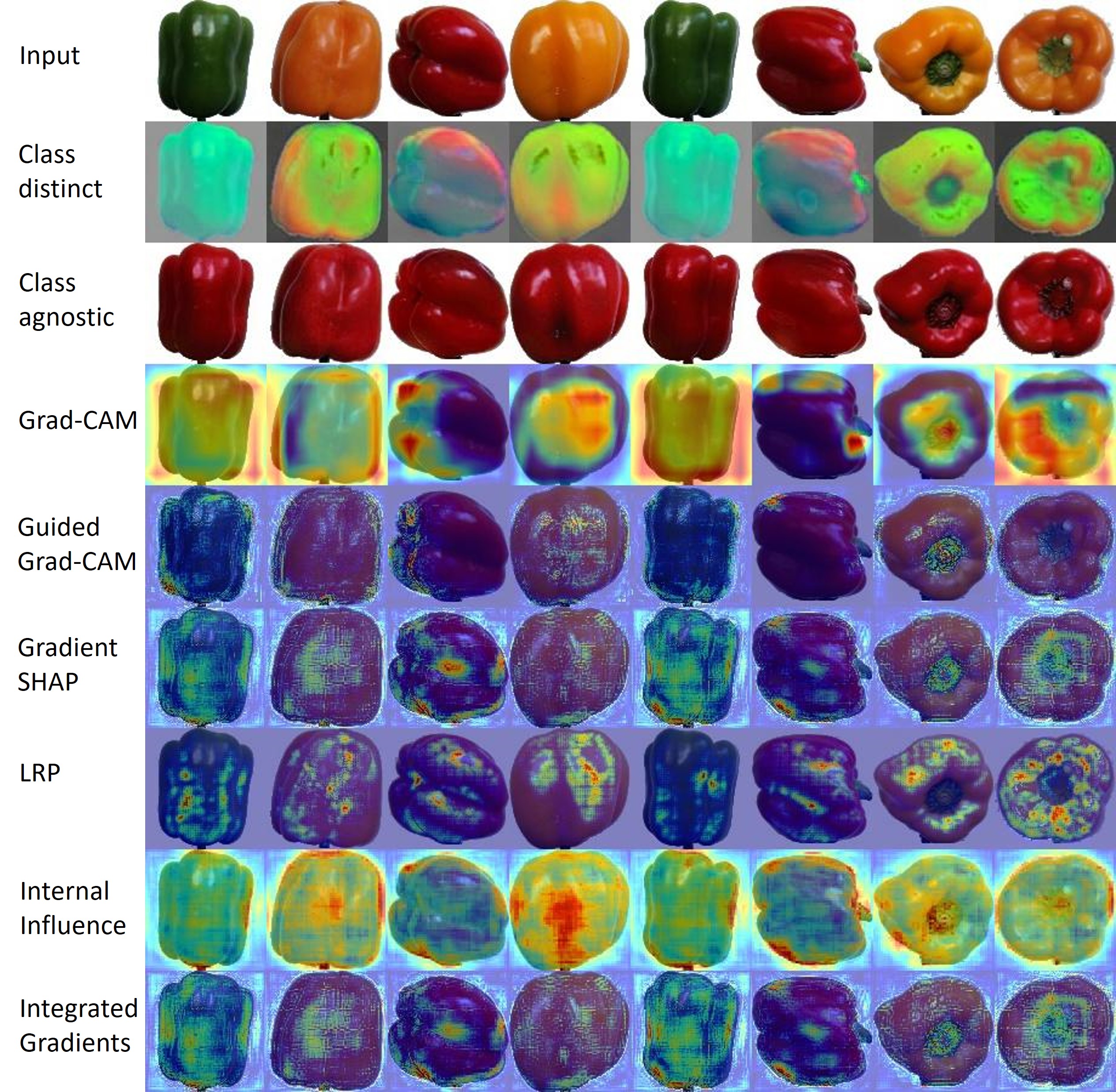}
    \caption{Peppers dataset ($C=D$). Class distinction is mainly by color, which is well visualized by our explanation method.}
    \label{fig:peppers_all_heatmaps_compare_disc}
\end{figure*}

\begin{figure*}[p]
    \centering
    \includegraphics[width=1\textwidth]{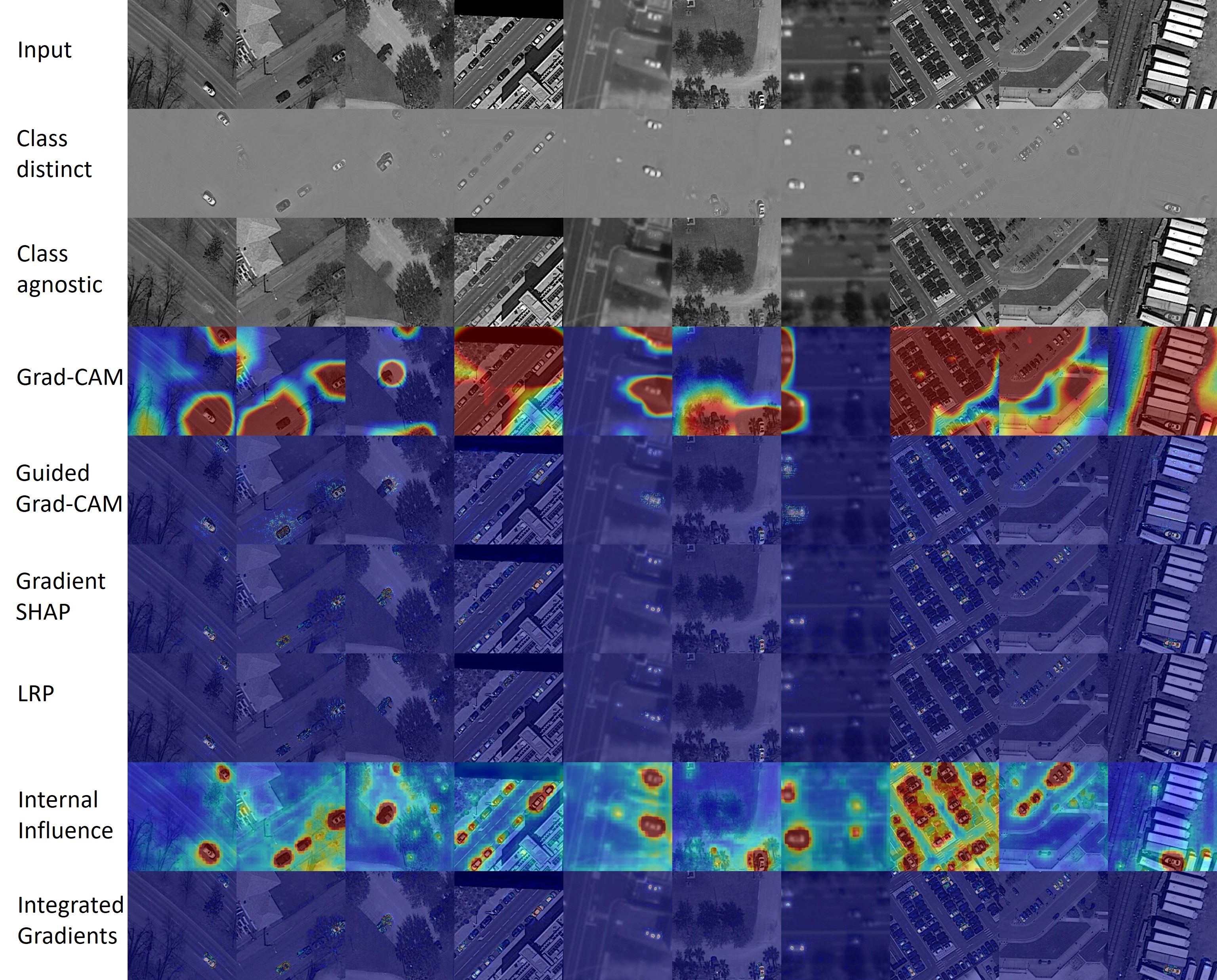}
    \caption{DOTA (cars) dataset ($C=D$). Cars are isolated well in the distinction part.}
    \label{fig:cars_all_heatmaps_compare_disc}
\end{figure*}

\begin{figure*}[p]
    \centering
    \includegraphics[width=1\textwidth]{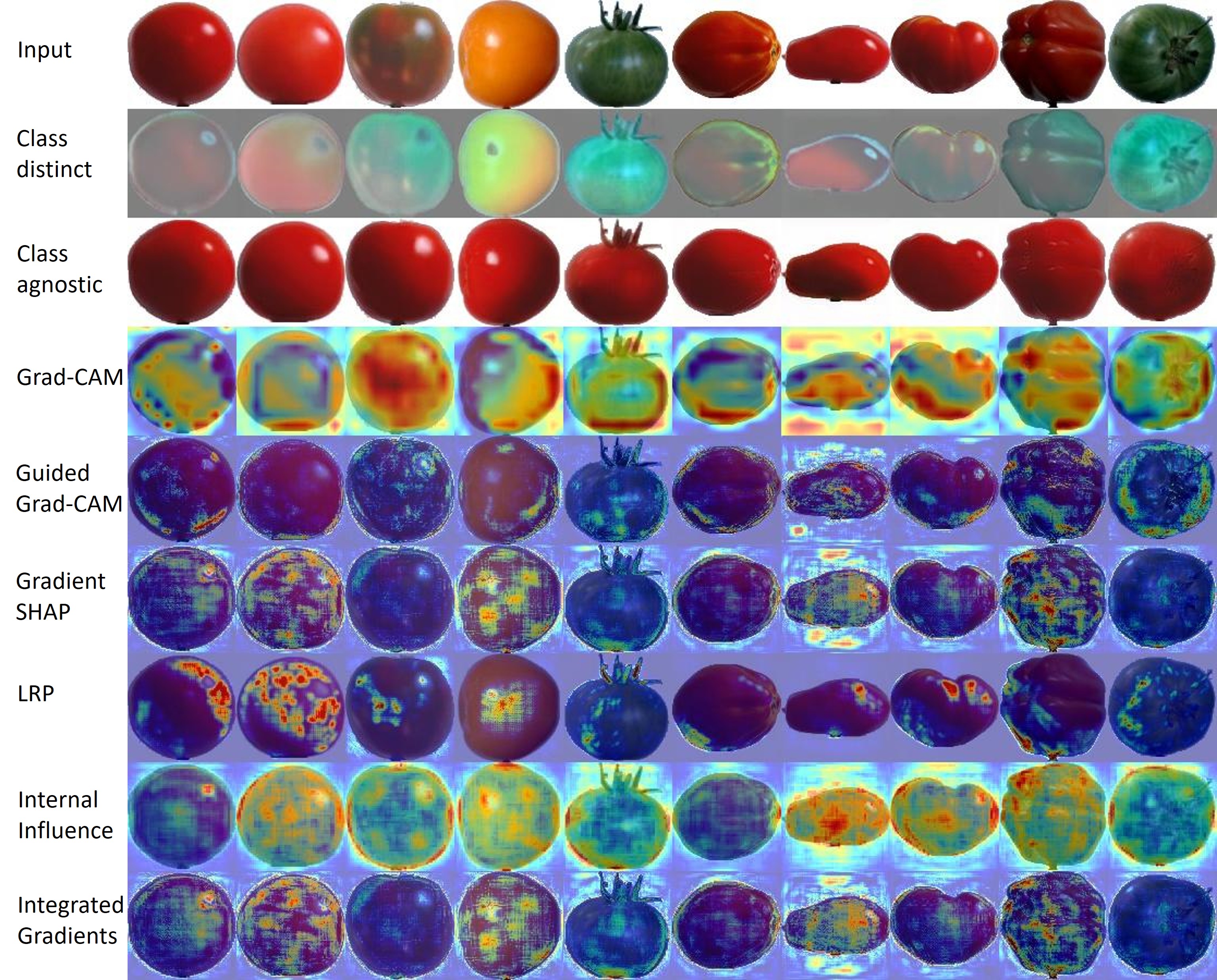}
    \caption{Tomatoes dataset ($C=D$). Distinction is mostly based on color (and some global texture).}
    \label{fig:tomatoes_all_heatmaps_compare_disc}
\end{figure*}

\begin{figure*}[p]
    \centering
    \includegraphics[width=1\textwidth]{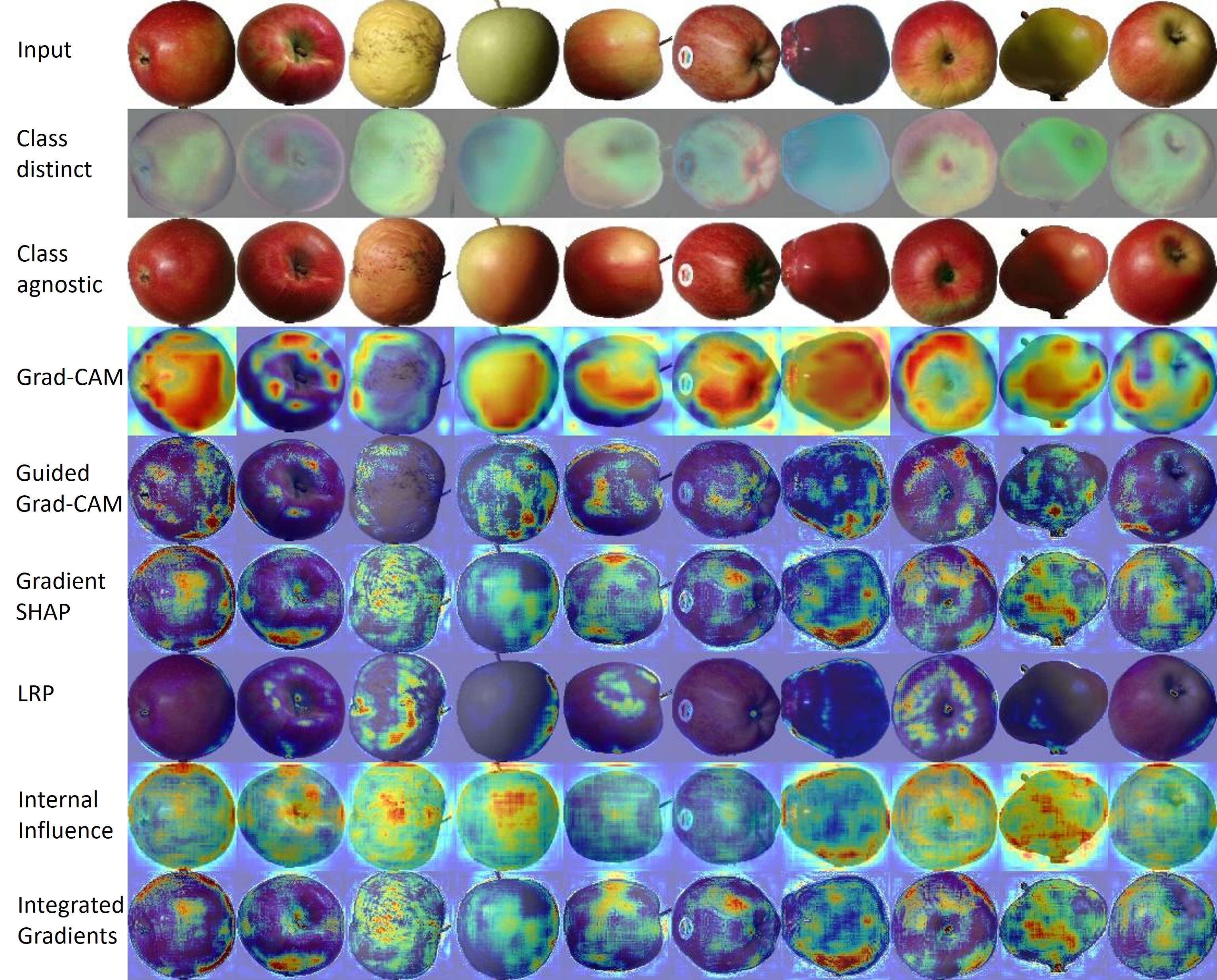}
    \caption{Apples dataset ($C=D$). Distinction by color is well captured by our methods.}
    \label{fig:apples_all_heatmaps_compare_disc}
\end{figure*}

\begin{figure*}[p]
    \centering
    \includegraphics[width=1\textwidth]{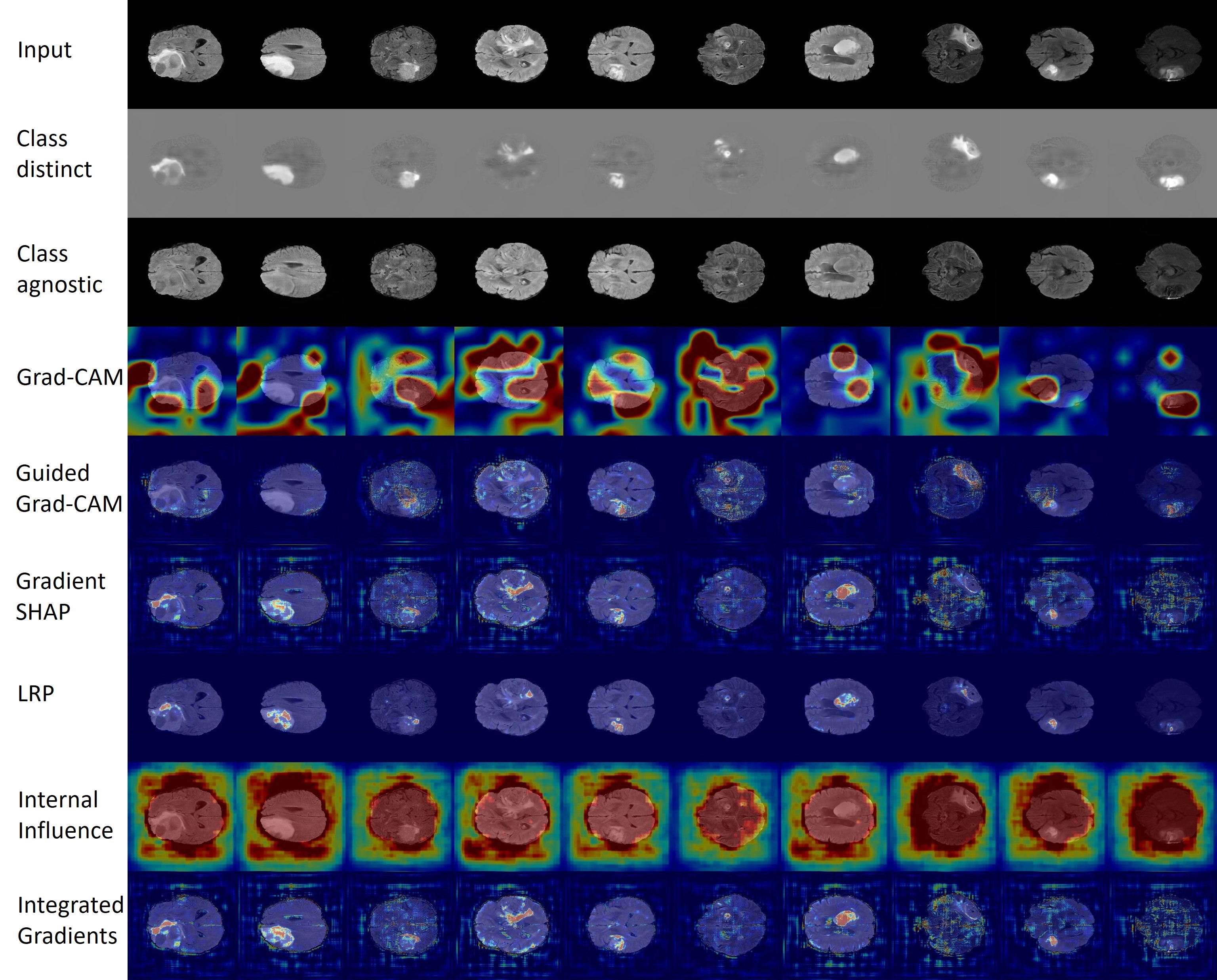}
    \caption{BraTS dataset ($C=D$). Highly localized regions of the lesions are visualized well in high resolution by our method.}
    \label{fig:brats_all_heatmaps_compare_disc}
\end{figure*}

\begin{figure*}[p]
    \centering
    \includegraphics[width=1\textwidth]{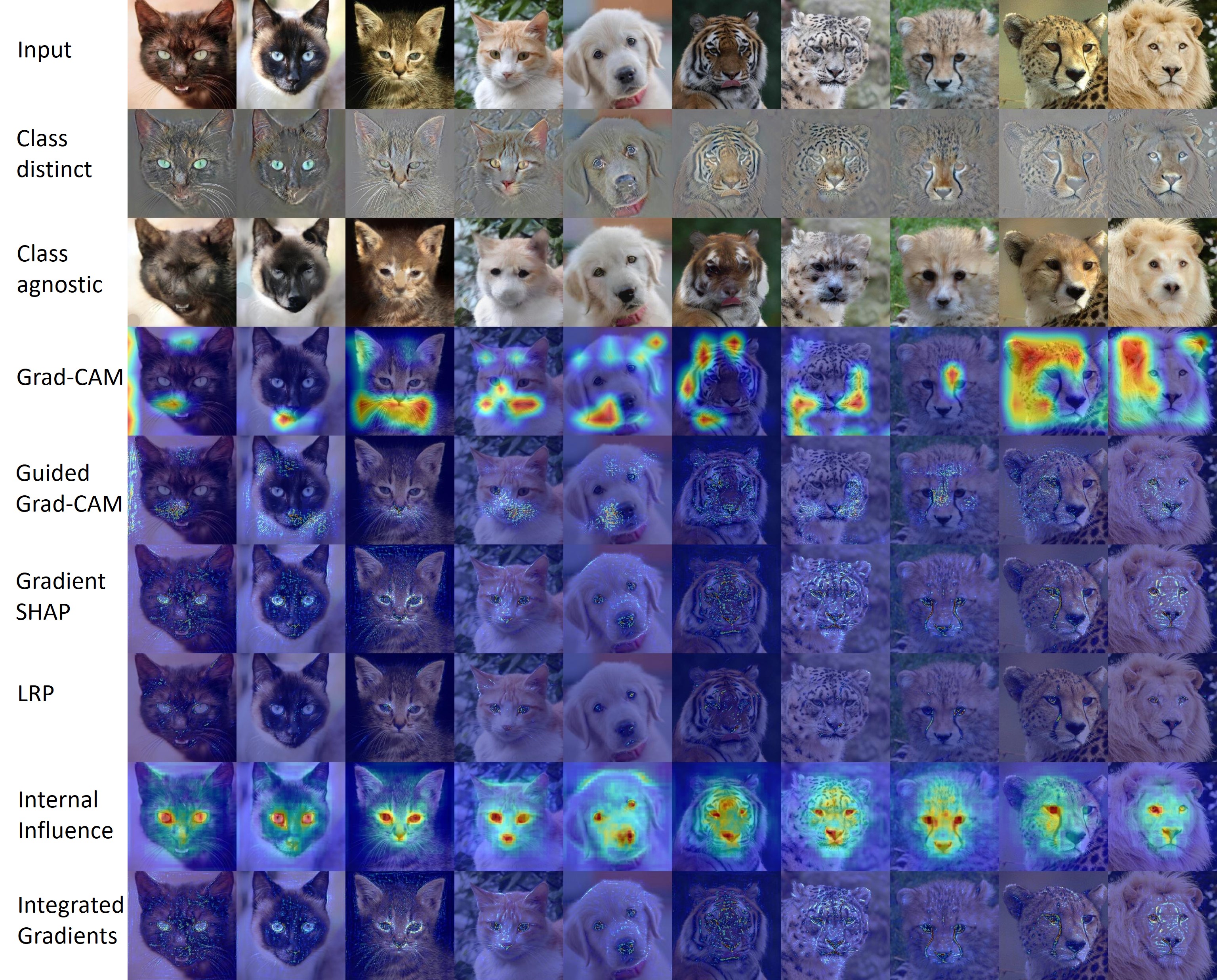}
    \caption{AFHQ animals dataset ($C=D$). Fur textures, as well as local features of eyes and nose give a full and rich explanation for class distinction. Some spurious background features appear, indicating background cues may contribute, in some cases, for the classifier's decision.}
    \label{fig:afhq_all_heatmaps_compare_disc}
\end{figure*}

\begin{figure*}[p]
    \centering
    \includegraphics[width=1\textwidth]{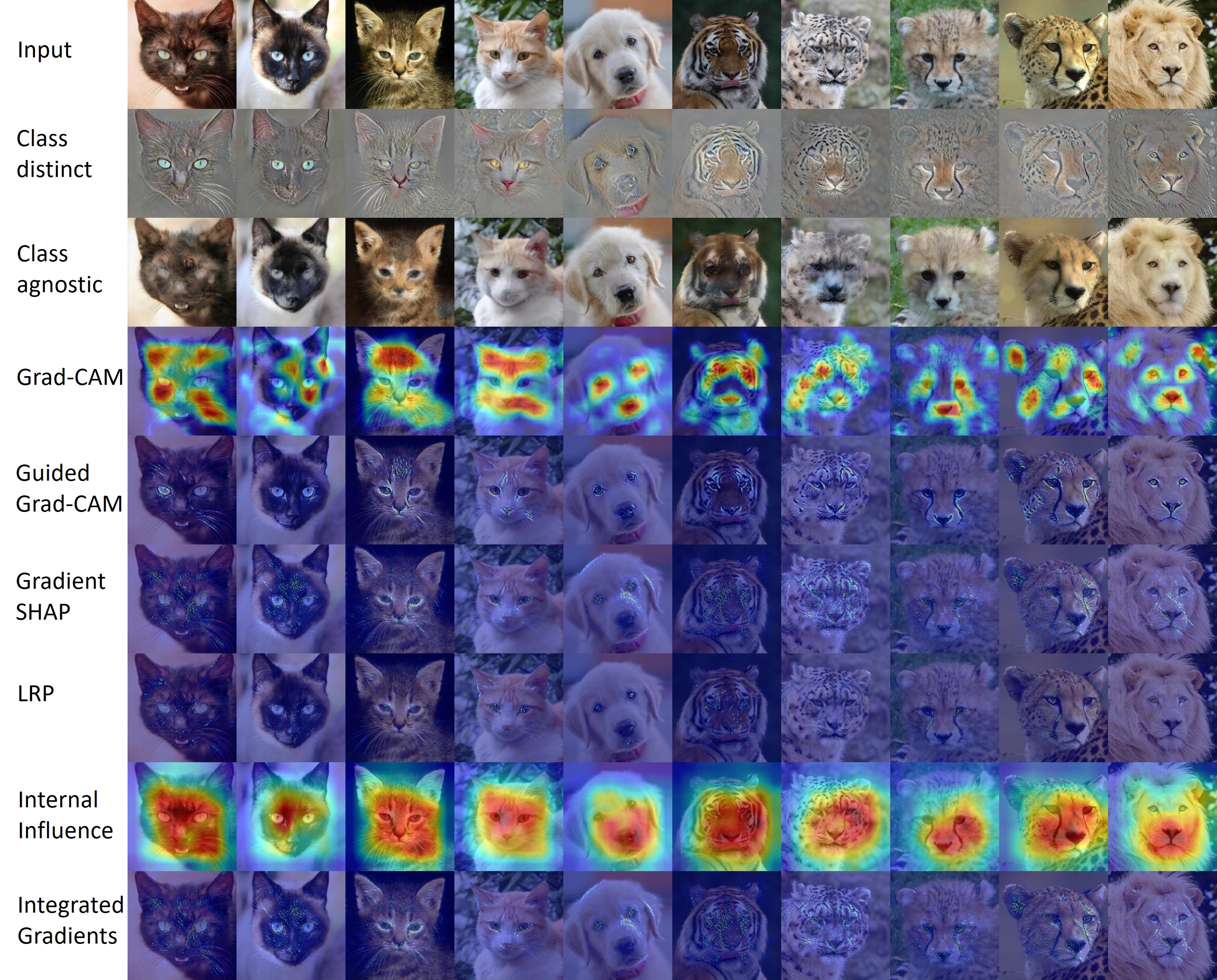}
    \caption{AFHQ animals dataset ($C=ResNet18$). Here ResNet18 is used as a classifier. Since both the discriminator and ResNet18 are highly accurate we can expect mostly similar distinct features. Whereas some XAI algorithm provide similar heatmaps, others are less stable. This depends also on meta-parameters, such as layer selection. Our algorithm shows only minor decomposition differences and appears stable.}
    \label{fig:afhq_all_heatmaps_compare_resnet18}
\end{figure*}

\begin{figure*}[p]
    \centering
    \includegraphics[width=1\textwidth]{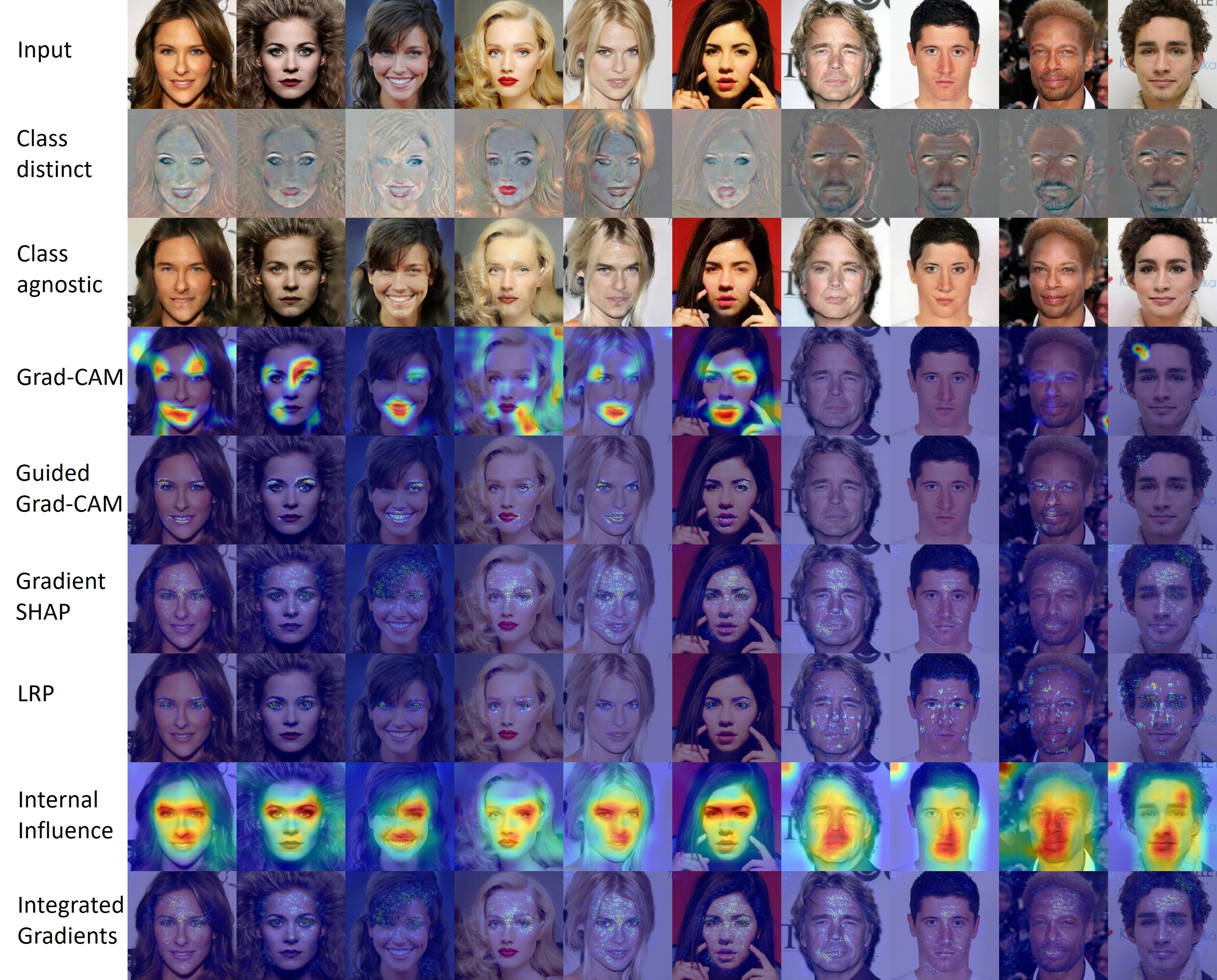}
    \caption{CelebA dataset ($C=ResNet18$).}
    \label{fig:celeb_all_heatmaps_compare_resnet18}
\end{figure*}

\begin{figure*}[p]
    \centering
    \includegraphics[width=1\textwidth]{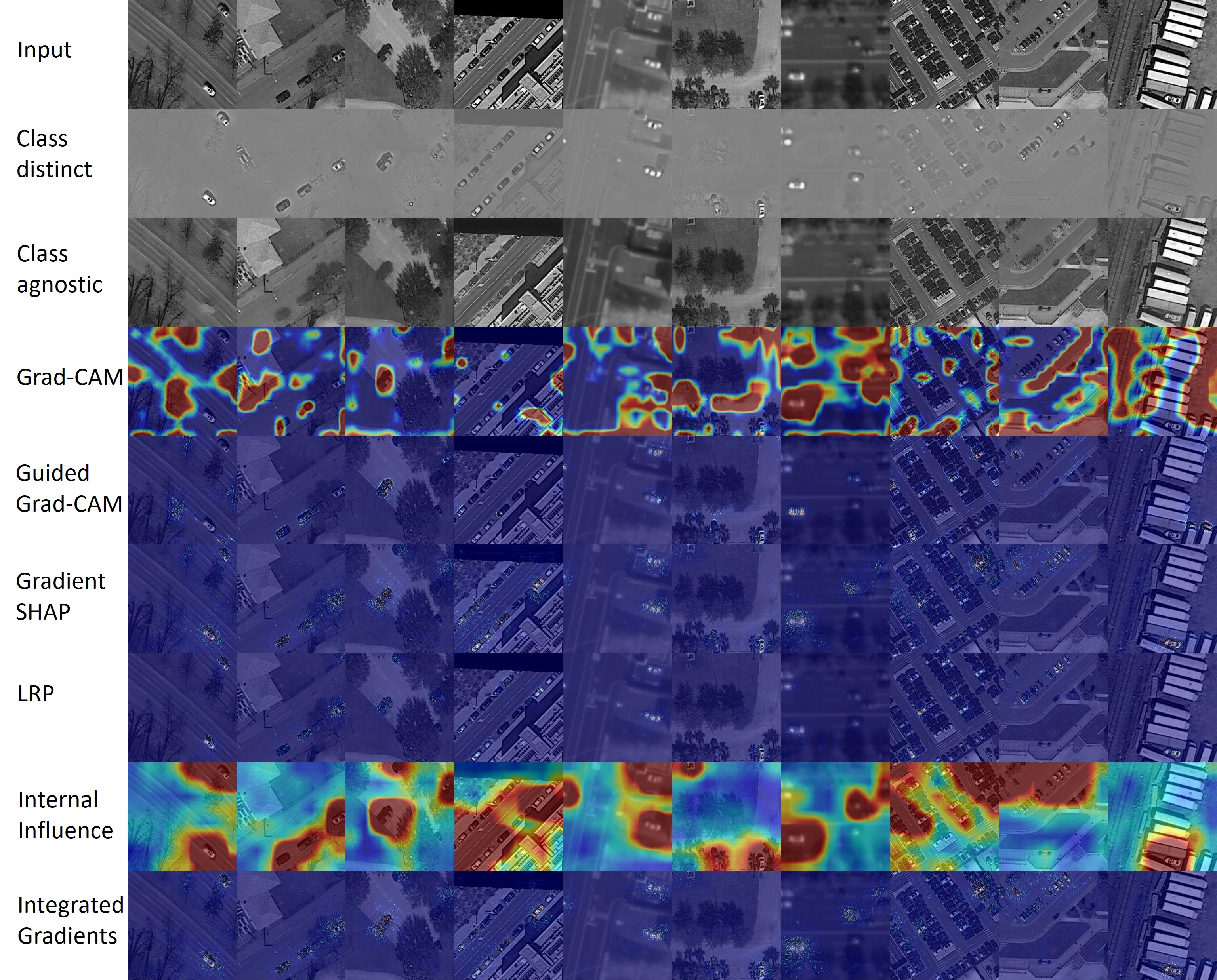}
    \caption{DOTA (cars) dataset ($C=ResNet18$).}
    \label{fig:cars_all_heatmaps_compare_resnet18}
\end{figure*}

\begin{figure*}[p]
\centering
\includegraphics[width=1\textwidth]{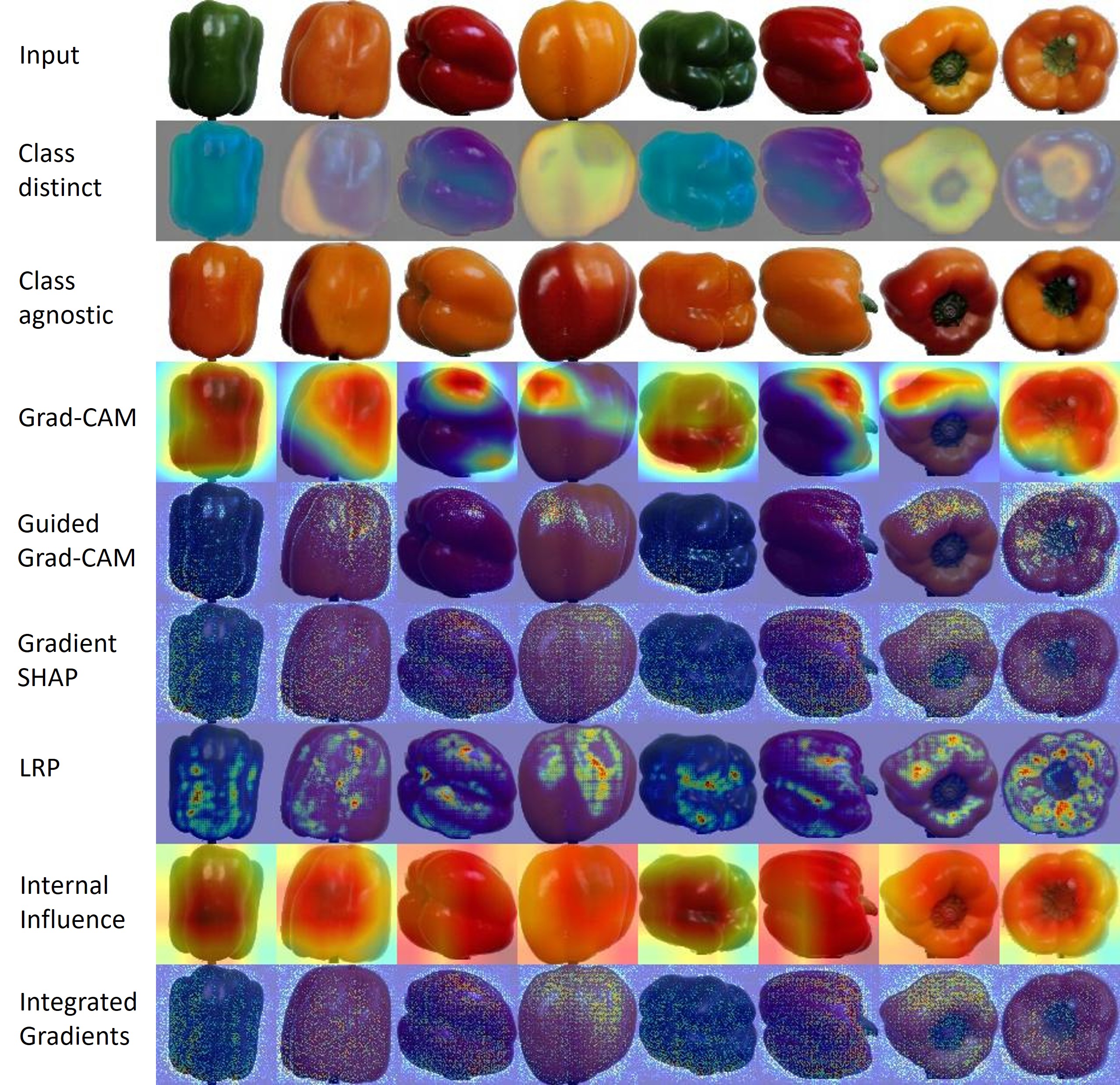}
\caption{Peppers dataset ($C=ResNet18$). Here we get different agnostic and distinction parts, compared to the case $C=D$. However, color is still the main distinctive feature and is well visualized, compared to heatmap methods.}
\label{fig:peppers_all_heatmaps_compare_resnet18}
\end{figure*}

\begin{figure*}[p]
    \centering
    \includegraphics[width=1\textwidth]{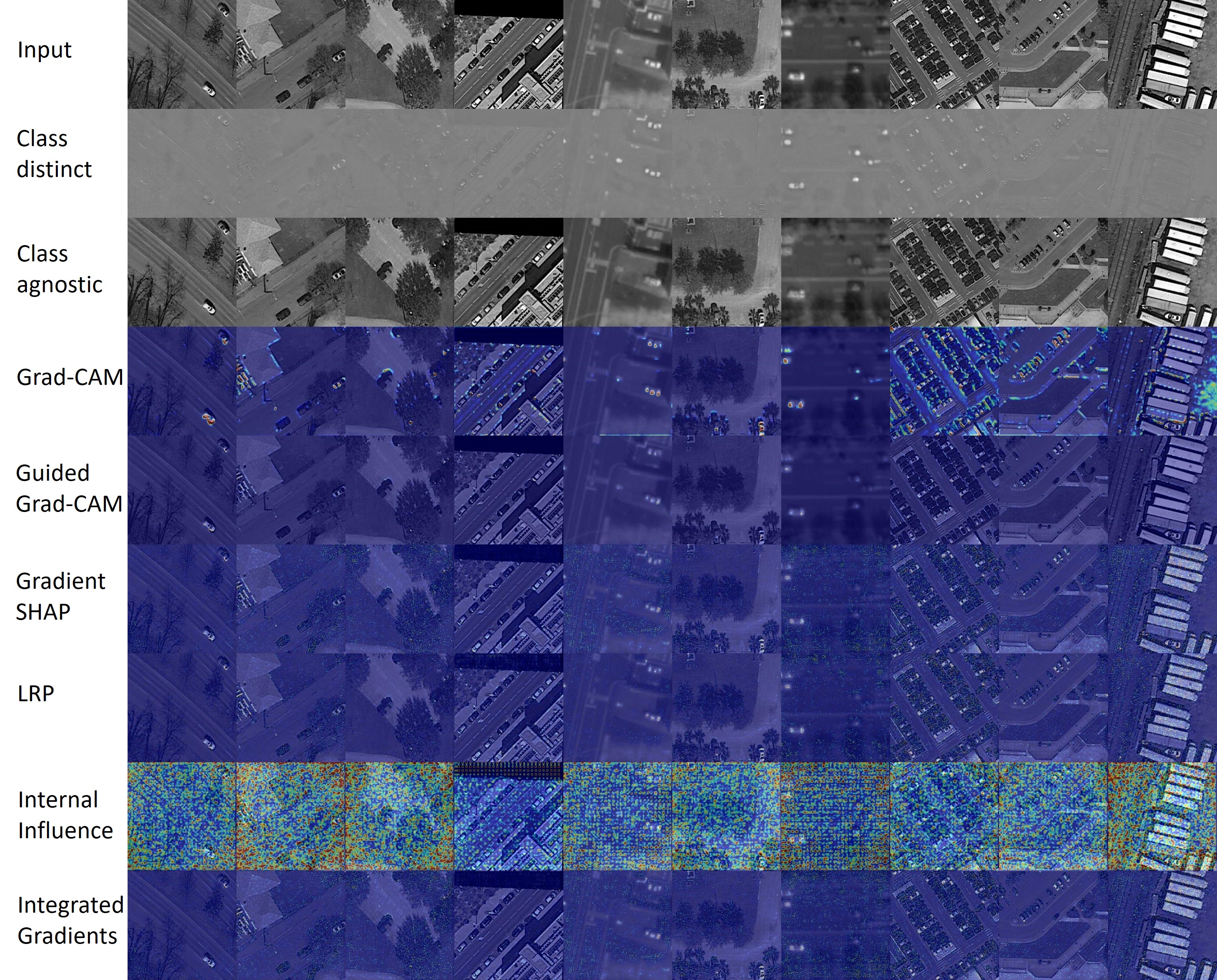}
    \caption{DOTA (cars) dataset ($C=Simple$). It can be observed that the distinction map is different and less clear compared to the maps generated by the discriminator and ResNet18. We hypothesize that this is due to the classifier's lower accuracy percentages, leading to a less precise differentiation between important features and those that are not.}
    \label{fig:cars_all_heatmaps_compare_simple}
\end{figure*}

\begin{figure*}[p]
    \centering
    \includegraphics[width=1\textwidth]{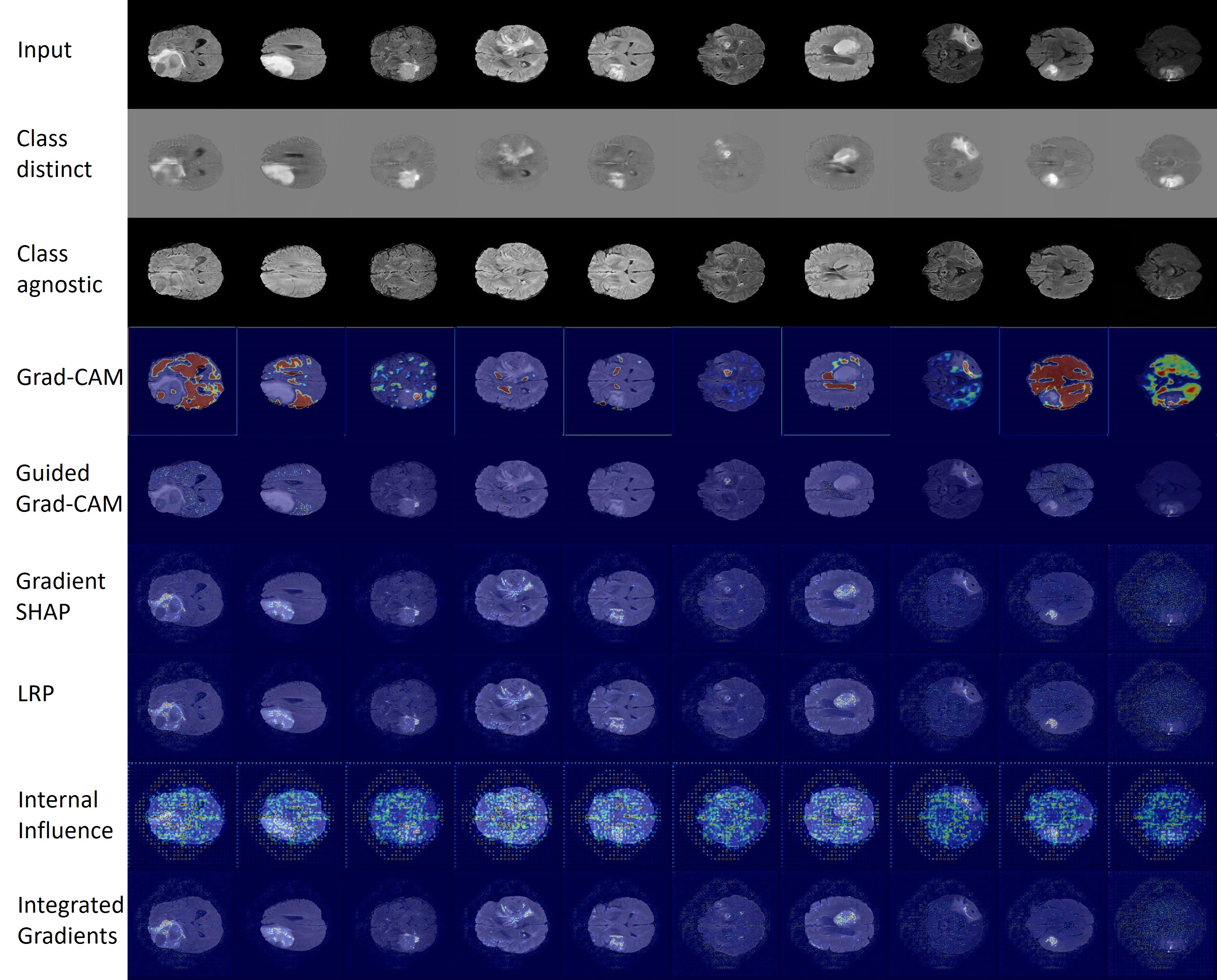}
    \caption{BraTS dataset ($C=Simple$). As in \cref{fig:cars_all_heatmaps_compare_simple}, here too, you can observe that the distinction maps are less clean, and we assume that this is due to similar reasons.}
    \label{fig:brats_all_heatmaps_compare_simple}
\end{figure*}

\begin{figure*}[p]
    \centering
    \includegraphics[width=1\textwidth]{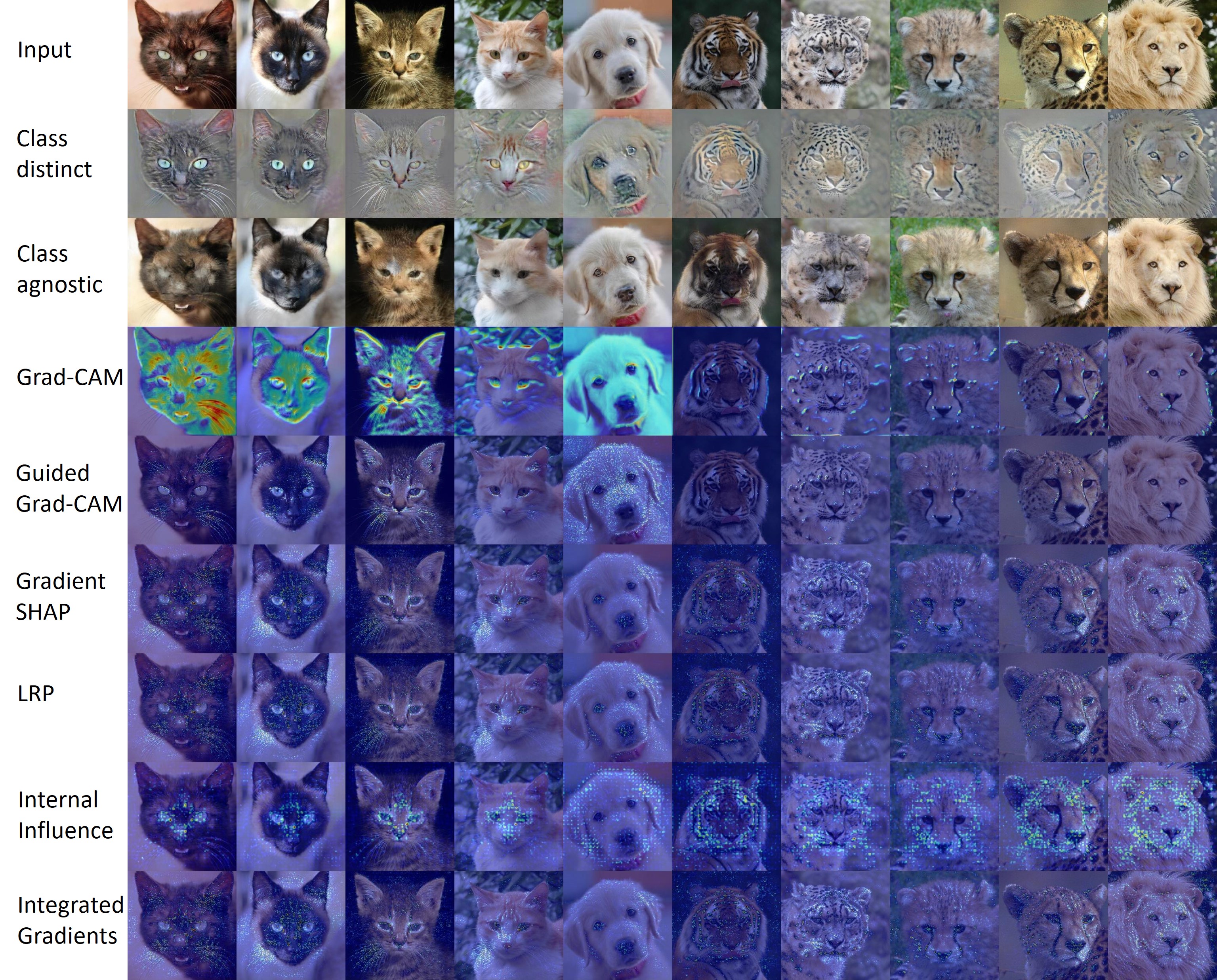}
    \caption{AFHQ dataset ($C=Simple$).}
    \label{fig:afhq_all_heatmaps_compare_simple}
\end{figure*}

\begin{figure*}[p]
    \centering
    \includegraphics[width=1\textwidth]{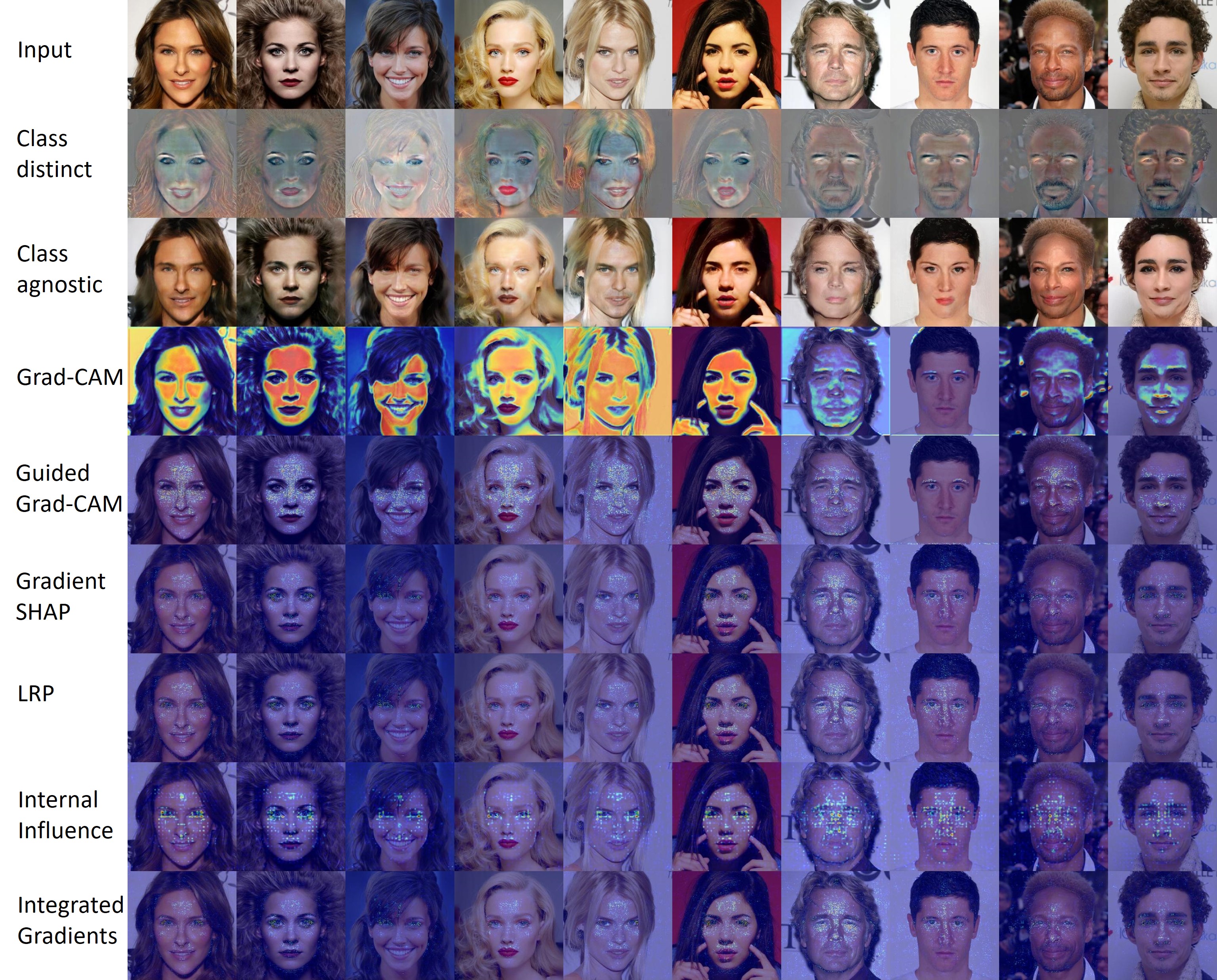}
    \caption{CelebA dataset ($C=Simple$).}
    \label{fig:celeba_all_heatmaps_compare_simple}
\end{figure*}

In the main paper we presented qualitative results of our algorithm for different datasets with selected comparisons to other methods.
Here we provide more details.
We show for each image the full comparison to 6 established XAI methods. We also provide more examples in each dataset, so the reader can judge better the quality and stability of our algorithm. 
We show the results for the case $C=D$ where the classifier is the discriminator (where classification accuracy is best) and for $C=ResNet18$, cases which were not shown in the main paper.
In \cref{fig:celeb_all_heatmaps_compare_disc}
we show comparison between Female and Male classes. There are many subtle cues for such distinction. One can observe that our method provides dense informative explanations which highlight well the distinctions.
In \cref{fig:peppers_all_heatmaps_compare_disc} the peppers are shown, which differ mainly by color, as our proposed method shows clearly. Heatmap visualization cannot show well global color explanations. 
In \cref{fig:cars_all_heatmaps_compare_disc} we show that the cars are well isolated in high resolution by our method. This case is less dense and can be handled well also by some other methods (such as LRP, Guided grad-cam or Gradient SHAP). Our method appears to visualize the explanation very clearly.
In \cref{fig:tomatoes_all_heatmaps_compare_disc} and 
\cref{fig:apples_all_heatmaps_compare_disc} we 
show additional datasets of tomatoes and apples classes, which did not appear in the main paper.
Here again color changes are a main feature of distinction, which is shown clearly by our method.
In \cref{fig:brats_all_heatmaps_compare_disc} additional
examples of the BraTS dataset are provided, showing our ability to isolate the lesions well. 
In \cref{fig:afhq_all_heatmaps_compare_disc} we observe the dense distinction images of animal classes, where different fur textures are significant features. These are very hard to visualize by heatmaps.
The above images were using the discriminator as the classifier. 
In \cref{fig:afhq_all_heatmaps_compare_resnet18},
\cref{fig:celeb_all_heatmaps_compare_resnet18},
\cref{fig:cars_all_heatmaps_compare_resnet18}
and
\cref{fig:peppers_all_heatmaps_compare_resnet18}
we show results for ResNet18 as the classifier.
We can expect that since both classifiers are relatively advanced, with high precision, they consider similar features. 
We can observe that our method produce qualitatively quite similar results and is stable. For the peppers case, \cref{fig:peppers_all_heatmaps_compare_resnet18}, we observe a stronger difference, where the 
agnostic part is more of a mix of orange and red colors.
In any case, the color distinction is clear and is much better visualized than in standard XAI methods.
In Figs \ref{fig:cars_all_heatmaps_compare_simple}, \ref{fig:brats_all_heatmaps_compare_simple}, \ref{fig:afhq_all_heatmaps_compare_simple}, \ref{fig:celeba_all_heatmaps_compare_simple} the classification explanations for the Simple classifier are shown. One can observe that the distinction map is less accurate, more
blurry, containing some spurious background features. This can be expected for a lower quality classifier.


\end{document}